%% file: main.tex
\documentclass{article}

\PassOptionsToPackage{linktoc=all}{hyperref}
\usepackage{jmlr2e}

\usepackage[utf8]{inputenc} 
\usepackage[T1]{fontenc}    
\usepackage{pifont}         
\usepackage{hyperref}       
\usepackage{url}            
\usepackage{booktabs}       
\usepackage{amsfonts}       
\usepackage{nicefrac}       
\usepackage{microtype}      
\usepackage[dvipsnames,hyperref,table]{xcolor}
\usepackage{amsmath}
\usepackage{amssymb}
\usepackage{subcaption}
\usepackage[font=small]{caption}
\usepackage{mathtools}
\usepackage{algorithm}
\usepackage{algorithmic}
\usepackage{multirow}
\usepackage{tabularx}
\usepackage{siunitx}
\usepackage{enumitem}
\usepackage{thmtools}
\usepackage{thm-restate}
\usepackage{xspace}
\usepackage{makecell}
\usepackage{adjustbox}
\usepackage{setspace}
\usepackage{placeins}
\usepackage{tocloft}
\usepackage{color, colortbl}
\definecolor{Gray}{gray}{0.93}

\newcolumntype{C}{>{\centering\arraybackslash}X}
\newcolumntype{L}{>{\raggedright\arraybackslash}X}
\newcolumntype{R}{>{\raggedleft\arraybackslash}X}
\input{std-macros}
\input{macros}

\newcommand\footnoteref[1]{\protected@xdef\@thefnmark{\ref{#1}}\@footnotemark}

\setlist[enumerate]{leftmargin=*}
\setlist[itemize]{leftmargin=*}

\expandafter\def\expandafter\UrlBreaks\expandafter{\UrlBreaks
  \do\a\do\b\do\c\do\d\do\e\do\f\do\g\do\h\do\i\do\j%
  \do\k\do\l\do\m\do\n\do\o\do\p\do\q\do\r\do\s\do\t%
  \do\u\do\v\do\w\do\x\do\y\do\z\do\A\do\B\do\C\do\D%
  \do\E\do\F\do\G\do\H\do\I\do\J\do\K\do\L\do\M\do\N%
  \do\O\do\P\do\Q\do\R\do\S\do\T\do\U\do\V\do\W\do\X%
  \do\Y\do\Z}

\title{Stronger Data Poisoning Attacks Break \\ Data Sanitization Defenses}

\author{%
  Pang Wei Koh\thanks{These authors contributed equally to this work.\\
  Published in \textit{Machine Learning}, 2021. Copyright 2021 by the authors.\\
  This paper was first published on arXiv in 2018 and has since been edited for clarity.}
      \email{pangwei@cs.stanford.edu}\\
      \addr{Computer Science Department\\Stanford University}
      \vspace{3mm}
  \AND Jacob Steinhardt\footnotemark[\value{footnote}]
      \email{jsteinhardt@berkeley.edu}\\
      \addr{Department of Statistics\\UC Berkeley}
      \vspace{3mm}
  \AND Percy Liang
      \email{pliang@cs.stanford.edu}\\
      \addr{Computer Science Department\\Stanford University}
}

\hypersetup{
  colorlinks   = true, 
  urlcolor     = NavyBlue, 
  linkcolor    = Blue, 
  citecolor    = NavyBlue 
}
\begin{document}

\date{}
\editor{}
\maketitle
\begin{abstract}
  \input abstract
\end{abstract}

\input intro
\input problem
\input attack
\input specific_attacks
\input experiments
\input transfer
\input related
\input discussion

\section*{Reproducibility}
Code and data for replicating our experiments are available at \url{https://github.com/kohpangwei/data-poisoning-journal-release}.

\section*{Acknowledgements}
We are grateful to Steve Mussmann, Zhenghao Chen, Marc Rasi, Robin Jia, and our anonymous reviewers for helpful comments and discussion. This work was partially funded by an Open Philanthropy Project Award. PWK was supported by the Facebook Fellowship Program.
JS was supported by the Fannie and John Hertz Foundation Fellowship.

\bibliography{refdb/all}

\appendix
\input svm-proof
\newpage
\input app_attack_details
\input experiments_mnist_dogfish
\newpage
\input alfa-details
\end{document}

%% file: std-macros.tex
\newcommand\sD{\ensuremath{\mathcal{D}}}

\newcommand\sF{\ensuremath{\mathcal{F}}}
\newcommand\sG{\ensuremath{\mathcal{G}}}

\newcommand\sL{\ensuremath{\mathcal{L}}}

\newcommand\sT{\ensuremath{\mathcal{T}}}

\newcommand\sX{\ensuremath{\mathcal{X}}}
\newcommand\sY{\ensuremath{\mathcal{Y}}}

\newcommand\bI{\ensuremath{\mathbf{I}}}

\newcommand\bR{\ensuremath{\mathbf{R}}}

\newcommand\bZ{\ensuremath{\mathbf{Z}}}

\DeclareMathOperator*{\sign}{sign}
\newcommand\p[1]{\ensuremath{\left( #1 \right)}} 
\newcommand\inv[1]{\ensuremath{\frac{1}{#1}}}

\newcommand\R{\ensuremath{\mathbb{R}}} 
\newcommand\Z{\ensuremath{\mathbb{Z}}} 
\newcommand\eqdef{\ensuremath{\stackrel{\rm def}{=}}} 
\newcommand\refeqn[1]{(\ref{eqn:#1})}

\newcommand\refsec[1]{Section~\ref{sec:#1}}
\newcommand\refsecs[2]{Sections~\ref{sec:#1} and~\ref{sec:#2}}
\newcommand\reffig[1]{Figure~\ref{fig:#1}}

\newcommand\reftab[1]{Table~\ref{tab:#1}}
\newcommand\refapp[1]{Appendix~\ref{sec:#1}}
\newcommand\refthm[1]{Theorem~\ref{thm:#1}}

\newcommand\reflem[1]{Lemma~\ref{lem:#1}}

\newcommand\refalg[1]{Algorithm~\ref{alg:#1}}

\newcommand\refprop[1]{Proposition~\ref{prop:#1}}

\newcommand\refcor[1]{Corollary~\ref{cor:#1}}

\ifthenelse{\isundefined{\definition}}{\newtheorem{definition}{Definition}}{}
\ifthenelse{\isundefined{\assumption}}{\newtheorem{assumption}{Assumption}}{}
\ifthenelse{\isundefined{\hypothesis}}{\newtheorem{hypothesis}{Hypothesis}}{}
\ifthenelse{\isundefined{\proposition}}{\newtheorem{proposition}{Proposition}}{}
\ifthenelse{\isundefined{\theorem}}{\newtheorem{theorem}{Theorem}}{}
\ifthenelse{\isundefined{\lemma}}{\newtheorem{lemma}{Lemma}}{}
\ifthenelse{\isundefined{\corollary}}{\newtheorem{corollary}{Corollary}}{}
\ifthenelse{\isundefined{\alg}}{\newtheorem{alg}{Algorithm}}{}
\ifthenelse{\isundefined{\example}}{\newtheorem{example}{Example}}{}
\newcommand{\E}{\ensuremath{\mathbb{E}}} 

%% file: macros.tex
\newcommand\mnist{MNIST-1-7}

\newcommand\Lerr{L_{\text{0-1}}}
\newcommand\sGt{\sG_{\hat\theta}}
\newcommand\sGty{\sG_{\hat\theta, y}}
\newcommand\sDc{\sD_\text{c}}
\newcommand\sDp{\sD_\text{p}}
\newcommand\tDp{\tilde{\sD}_\text{p}}
\newcommand\sDtest{\sD_\text{test}}
\newcommand\sDdecoy{\sD_\text{decoy}}
\newcommand\sDho{\sD_\text{pool}}
\newcommand\sDflip{\sD_\text{flip}}

\newcommand\sDcp{\sDc \cup \sDp}

\newcommand\gtest{g_{\hat\theta, \sDtest}}
\newcommand\gDc{g_{\thetadecoy, \sDc}}

\newcommand\sFb{\sF_{\beta}}

\newcommand\sFy{\sF_{y}}
\newcommand{\T}[1]{\theta_{#1}}

\DeclareMathOperator*{\argmax}{argmax}
\DeclareMathOperator*{\argmin}{argmin}

\newcommand\car{Carath\'eodory}

\newcommand{\xt}{\tilde{x}}
\newcommand{\yt}{\tilde{y}}
\newcommand{\ntest}{n_{\mathrm{test}}}
\newcommand{\nho}{n_{\mathrm{pool}}}
\newcommand{\sDsan}{\mathcal{D}_{\text{san}}}
\newcommand{\Ltwo}{{L2}}
\newcommand{\slab}{{slab}}
\newcommand{\Loss}{{loss}}
\newcommand{\knn}{{k-NN}}
\newcommand{\PCA}{{SVD}}
\newcommand{\alfa}{{alfa}}
\newcommand{\infbasic}{{influence-basic}}
\newcommand{\mmbasic}{{min-max-basic}}
\renewcommand{\inf}{{influence}}
\newcommand{\mm}{{min-max}}
\newcommand{\kkt}{{KKT}}
\newcommand{\xp}{\tilde{x}}
\newcommand{\xpp}{\tilde{x}_{+}}
\newcommand{\xpn}{\tilde{x}_{-}}
\newcommand{\epp}{\epsilon_{+}}
\newcommand{\epn}{\epsilon_{-}}
\newcommand{\yp}{\tilde{y}}
\newcommand{\ceil}[1]{\lceil #1 \rceil}
\newcommand{\floor}[1]{\lfloor #1 \rfloor}

\newcommand{\nb}{n_{\text{burn}}}
\newcommand{\thetac}{\theta_c}
\newcommand{\thetadecoy}{\theta_{\text{decoy}}}

%% file: abstract.tex
Machine learning models trained on data from the outside world can be corrupted by
\emph{data poisoning} attacks that inject malicious points into the models' training sets.
A common defense against these attacks is \emph{data sanitization}:
first filter out anomalous training points before training the model.
In this paper, we develop three attacks that can bypass a broad range of common data sanitization defenses,
including anomaly detectors based on nearest neighbors, training loss, and singular-value decomposition.
By adding just 3\% poisoned data, our attacks successfully increase test error on the Enron spam detection dataset from 3\% to 24\% and on the IMDB sentiment classification dataset from 12\% to 29\%.
In contrast, existing attacks which do not explicitly account for these data sanitization defenses are defeated by them.
Our attacks are based on two ideas:
(i) we coordinate our attacks to place poisoned points near one another, and
(ii) we formulate each attack as a constrained optimization problem, with constraints designed to ensure that the poisoned points evade detection. As this optimization involves solving an expensive bilevel problem,
our three attacks correspond to different ways of approximating this problem, based on influence functions; minimax duality; and the Karush-Kuhn-Tucker (KKT) conditions.
Our results underscore the need to develop more robust defenses against data poisoning attacks.

%% file: intro.tex
\section{Introduction}\label{sec:intro}

\begin{figure}[h]
\begin{center}
\includegraphics[width=\columnwidth]{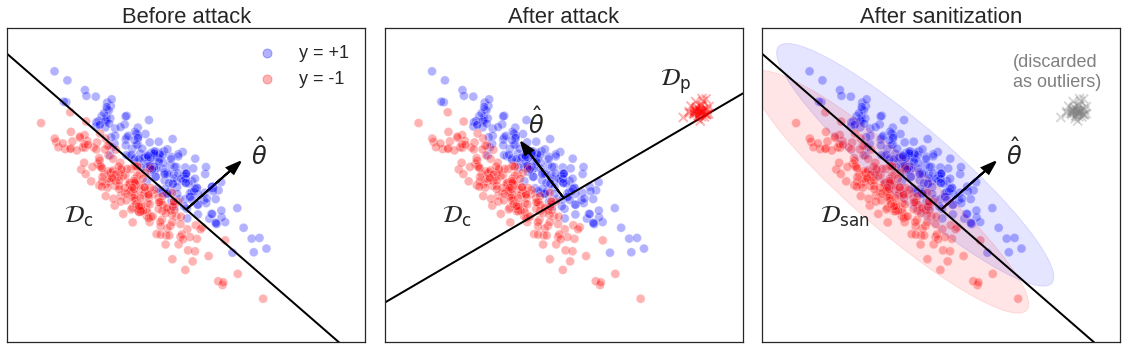}
\end{center}
\vspace{-5mm}
\caption{
{\bf Left:} In the absence of any poisoned data, the defender can often learn model parameters $\hat\theta$ that fit the true data $\sDc$ well.
Here, we show the decision boundary learned by a linear support vector machine on synthetic data.
{\bf Middle:} However, the addition of poisoned data $\sDp$ can significantly change the learned $\hat\theta$,
leading to high test error $\sL(\hat\theta)$.
{\bf Right:} By discarding outliers from $\sD = \sDcp$ and then training on the remaining $\sDsan$, the defender
can mitigate the effectiveness of the attacker. In this example, the defender discards all blue points outside the blue ellipse, and all red points outside the red ellipse.
}
\label{fig:setup}
\end{figure}

In high-stakes settings like autonomous driving \citep{gu2017badnets},
biometrics \citep{chen2017targeted}, and cybersecurity \citep{rubinstein2009antidote, suciu2018does},
machine learning (ML) systems need to be secure against attacks by malicious actors.
Securing ML systems is complicated by the fact that they are often trained on data obtained from the outside world,
which makes them especially vulnerable.
By attacking this data collection process, which could be as easy as creating a new user account,
adversaries can inject malicious data into the system and cause it to fail.

These \emph{data poisoning} attacks are the focus of the present work.
We consider attacks against classifiers where
an attacker adds a small fraction of new training points to
degrade the performance of the trained classifier on a test set.
\reffig{setup} illustrates this setting:
a model that might otherwise correctly classify most of the data (\reffig{setup}-Left)
can be made to learn a significantly different decision boundary
by an attacker who injects just a small amount of poisoned data (\reffig{setup}-Middle).

A common and often effective defense against data poisoning attacks is \emph{data sanitization},
which uses anomaly detectors to filter out suspicious training points \citep{hodge2004survey, cretu2008casting, paudice2018detection}.
\reffig{setup}-Right illustrates a hypothetical defense: by removing the anomalous poisoned data, the defender can learn the correct decision boundary.
In our experiments, data sanitization defenses were able to defeat all of the existing data poisoning attacks we tested.

However, those attacks were naive in that they did not explicitly try to evade the data sanitization defenses.
This is typical in the literature:
attackers might be optimized to act within an attack budget \citep{mei2015teaching}
and to only add points that belong to the input domain (e.g., word counts in a document should be integer-valued \citep{nelson2008exploiting, newell2014practicality},
but not to evade defenses.
Previous work has suggested that attacks optimized for evading data sanitization can in fact evade some types of defenses \citep{steinhardt2017certified},
whereas attacks that are not optimized can be easily detected as anomalies \citep{frederickson2018attack}.
This leads to the question of whether data sanitization defenses are vulnerable to attackers who explicitly try to evade anomaly detection.

In this paper, we answer this question in the affirmative.
We develop three data poisoning attacks that can simultaneously evade a broad range of common data sanitization defenses, including anomaly detectors based on nearest neighbors, training loss, singular-value decomposition, and the distance to class centroids.
Our attacks are also able to deal with integer constraints on the input, which naturally arise in domains like natural language.
For example, our attacks on a linear support vector machine increase test error on the Enron spam detection dataset from $3\%$ to $24\%$ and on the IMDB sentiment classification dataset from $12\%$ to $29\%$ by adding just $3\%$ poisoned data, even in the presence of these data sanitization defenses.

We adopt two strategies to evade data sanitization defenses.
The first strategy is targeted at defenses which use anomaly detectors that are highly sensitive to the presence of just a few points:
e.g., an anomaly detector that filters out points which are far from their nearest neighbors will not recognize a point as anomalous if it is surrounded by a few other points, even if those points is far from the rest of the data.
Intuitively, such detectors tend to `overfit' the training data.
To evade these defenses,
our attacks \emph{concentrate} poisoned points in just a few distinct locations (\refsec{attack}).
For example, poisoned data placed in a tight cluster will evade the nearest-neighbor-based anomaly detector that throws out points far away from other points.
We show theoretically that we can concentrate all of the attack mass on just a few distinct points (e.g., only 2 points for 2-class support vector machines (SVMs) and logistic regression models) without any loss in attack effectiveness.

The second strategy is targeted at defenses which use anomaly detectors that are less sensitive and highly parametric:
e.g., an anomaly detector that throws out points beyond some distance from the data centroid will not depend too much on the addition or removal of a few points from the data, as long as the data centroid does not change significantly.
These defenses are more resistant to concentrated attacks, since they are less sensitive to small changes in the data (in this paper, we consider only attacks that inject a small fraction---3\% or less---of poisoned data).
To evade them, we
formulate the attack as a constrained optimization problem,
where the objective is to maximize the test loss of the model that the defender learns on the union of the clean and poisoned data;
the optimization variables are the locations of the poisoned points;
and the constraints are imposed by the defenses (such that a point that satisfies the constraints will be guaranteed to evade the defenses).

Formulating a data poisoning attack as an optimization problem is a common technique, first introduced in the context of support vector machines in \citet{biggio2012poisoning} and subsequently refined by \citet{mei2015teaching} and later work.
The central difficulty is that the bilevel problem---the attacker needs to reason about what model the defender would learn, which in turn requires solving an inner optimization problem---is non-convex and intractable to solve exactly \citep{bard1991some},
and even local methods like gradient ascent can be slow \citep{koh2017understanding}.
This is made even more challenging in our setting, compared to prior work,
because we have additional constraints that encode the defenses that the attacker wishes to evade.
To overcome this computational hurdle, we use two ideas:
\begin{enumerate}
  \item We \textbf{concentrate the attacks} on a small number of distinct points. Beyond allowing us to evade some data sanitization defenses, as discussed above, it also significantly improves computational efficiency as we only need to optimize for the locations of a few distinct points. We use this to speed up gradient ascent on the bilevel optimization problem, leading to what we call the \inf{} attack (\refsec{improved-inf}).

  \item The bilevel problem is expensive to solve because the effect of the optimization variables (poisoned points) on the objective (test loss) depends on the model parameters that the defender learns on the poisoned data, an intermediate quantity that is expensive to compute.
  We break this dependency by first finding \textbf{decoy parameters}---model parameters that have high test error but low training error. Given fixed decoy parameters, we can then efficiently find poisoned points that yield the decoy parameters when trained on. We call this the \kkt{} attack (\refsec{kkt}), after the Karush-Kuhn-Tucker optimality conditions that used to derive this attack.
  Furthermore, we use these decoy parameters to adapt the attack introduced by \citet{steinhardt2017certified}, which in its original form is efficient but cannot evade loss-based defenses; this leads to our \mm{} attack (\refsec{improved-mm}).
  These attacks mitigate some of the drawbacks of the \inf{} attack, which is still too computationally intensive to scale to large datasets and sometimes gets stuck in local minima.
\end{enumerate}
Finally, our study reveals several surprising facts about data poisoning:
\begin{enumerate}
\item Poisoned data does not have to look anomalous; if the poisoned points are carefully coordinated, each poisoned point can appear normal, as in the above example of the nearest-neighbor based anomaly detector.

  \item Poisoned points need not have high loss under the poisoned model, so the defender cannot simply throw out points that have high loss. For example, given fixed decoy parameters, we can constrain our poisoned data to have low loss under those parameters.

  \item Regularization reduces the effect that any single data point can have on the model and is therefore tempting to use as a defense against data poisoning. However, increasing regularization can actually make the defender more susceptible to attacks, because the defender becomes less able to fit the small fraction of poisoned points.
\end{enumerate}

The success of our data poisoning attacks against common anomaly-based data sanitization defenses suggest that more work needs to be done on defending against data poisoning attacks. In particular, while anomaly detectors are well-suited to detect independently-generated anomalous points (e.g., due to some noise process in nature), a robust defense against data poisoning attacks will have to account for the ability of the attacker to place all of their poisoned data points in a coordinated fashion.

Beyond the merits of our specific attacks, our results underscore a broader point: data poisoning attacks need to account for defenses,
and defenses correspondingly need to account for attacks that are specifically targeted against them.
Attacks that do not consider defenses might work against a naive defender but be easily defeated by other defenses.
Similarly, defenses that are only tested against basic attacks might give a false sense of security, as they might be broken by more determined and coordinated attackers.

%% file: problem.tex
\section{Problem Setting and Defenses}\label{sec:problem}

\subsection{General setting}
\label{sec:general-setting}

We consider classification tasks. To simplify exposition, we focus on binary tasks, where we seek to learn a model $f_{\theta} : \sX \to \{-1,+1\}$,
parametrized by $\theta \in \R^d$, that maps from features
$x \in \sX$ to an output $y \in \{-1,+1\}$.
A model $f_{\theta}$ is evaluated on a fixed test set $\sDtest = \{(x_i, y_i)\}_{i=1}^{n_\text{test}}$ by its 0-1 test error $\Lerr(\theta; \sDtest)$,
which is the proportion of $\sDtest$ that it classifies wrongly:
\begin{equation}
\Lerr(\theta; \sDtest) = \frac{1}{|\sDtest|} \sum_{(x,y) \in \sDtest} \bI[f_{\theta}(x)\neq y].
\end{equation}

In the setting we consider, the \emph{defender} aims to pick a $\hat\theta$ with low test error $\Lerr(\hat\theta; \sDtest)$,
while the \emph{attacker} aims to mislead the defender into picking a $\hat\theta$ with high $\Lerr(\hat\theta; \sDtest)$.
The attacker observes the test set $\sDtest$ as well as a clean training set $\sDc = \{(x_i, y_i)\}_{i=1}^n$, and chooses $\epsilon n$ \emph{poisoned} points $\sDp$ to add to $\sDc$.
The defender observes the combined training set $\sD = \sDc \cup \sDp$ consisting of the original $n$ clean points
and the $\epsilon n$ additional poisoned points; uses a data sanitization defense to remove anomalous points; and then learns $\hat\theta$ from the remaining data.

The attacker has several advantages: it knows the test set in advance (whereas the defender does not);
it knows the defender's training procedure; and it also gets to observe the clean training set $\sDc$.
In reality, the attacker might not have access to all of this information.
However, as defenders, we want to be robust even to attackers that might have the above information (this is the principle of security by design; see, e.g., \citet{biggio2014security}).
For example, an attacker whose goal is to make the defender get a particular set of predictions wrong (e.g., the attacker might want to cause a ``fake news'' classifier to classify all websites from a certain domain as ``real news'') would accordingly choose, and therefore get to observe, $\sDtest$.
In contrast, the defender might not know the attacker's goal in advance, and therefore would not have access to $\sDtest$.

\noindent\fbox{%
    \parbox{\textwidth}{%
    \vspace{-2mm}
\paragraph{Attacker:} \begin{itemize}
\item Input: Clean training data $\sDc$ and test data $\sDtest$.
\item Output: Poisoned training data $\sDp$, with $|\sDp| = \epsilon |\sDc|$.
\item Goal: Mislead defender into learning parameters $\hat\theta$ with high test error $\Lerr(\hat\theta; \sDtest)$.
\end{itemize}
\paragraph{Defender:}
\begin{itemize}
\item Input: Combined training data $\sD = \sDc \cup \sDp$.
\item Output: Model parameters $\hat\theta$.
\item Goal: Learn model parameters $\hat\theta$ with low test error $\Lerr(\hat\theta; \sDtest)$ by filtering out poisoned points $\sDp$.
\end{itemize}
    }%
}
\vspace{3mm}

\paragraph{Further assumptions in experiments.}
In our experiments, we assume that $f_{\theta}$ is a linear classifier, i.e., $f_\theta(x) = \sign(\theta^\top x)$ for binary tasks. We consider both binary and multi-class classification.
We also focus on \emph{indiscriminate attacks} \citep{barreno2010security}, where the test data $\sDtest$ is drawn from the same distribution as the clean training data $\sDc$, and the attacker tries to increase the average test error of the defender's model under this underlying data distribution. In \refsec{transfer-train}, we show that the attacker can still construct strong indiscriminate attacks even when they do not know the test data $\sDtest$. Most of our methods are also applicable to more general choices of $\sDtest$; we discuss this further in \refsec{related}.

\subsection{Data sanitization defenses}\label{sec:defenses}

To thwart the attacker, we assume the defender employs a
\emph{data sanitization} defense
\citep{cretu2008casting}, which first removes anomalous-looking points from
$\sD = \sDc \cup \sDp$ and then trains on the remaining points.
The motivation is that intuitively, poisoned data that looks similar to the clean data will not be effective in changing the learned model;
therefore, the attacker would want to place poisoned points that are somehow different from the clean data.
By discarding points that look too different (anomalous), the defender can thus protect itself against attack.
Ideally---as in the hypothetical \reffig{setup}---a defense would discard the poisoned data $\sDp$ and leave the clean data $\sDc$, so that the defender learns model parameters $\hat\theta$ that have low test error.

Defenses differ in how they judge points as being anomalous.
To formalize this, we represent each defense by a \emph{score function} $s_\beta : \sX \times \sY \to \mathbb{R}$
that takes in a data point $(x, y)$ and returns a number representing
how anomalous that data point is.
This score function is parametrized by \emph{anomaly detector parameters} $\beta$ that are derived from the combination of the clean and poisoned training data $\sD = \sDc \cup \sDp$.
As an example, in what we call the \Ltwo{} defense, the defender discards points in $\sD$ that are far from their respective class centroids.
For this defense, $\beta = (\mu_+, \mu_-)$ would represent the class centroids in $\sD$ and $s_\beta(x, y)= \|x - \beta_y\|_2$ would measure the distance of $x$ to the centroid of class $y$.

Concretely, a defender $(B, s_\beta)$:
\begin{enumerate}
\item Fits the \emph{anomaly detector parameters} $\beta = B(\sD)$,
where $B$ is a function that takes in a dataset and returns a vector.

\item Constructs the \emph{feasible set} $\sF_\beta  = \{(x, y) : (x, y) \in \sX \times \sY \text{ with } s_\beta(x, y) < \tau_y\}$.
The threshold $\tau_y$ is chosen such that a desired fraction of points from each class are discarded.

\item Forms the sanitized training dataset $\sDsan = \sD \cap \sF_\beta$ by discarding all points that fall outside the feasible set.

\item Finds the $\hat\theta$ that minimizes the regularized training loss on $\sDsan$:
\begin{equation}
\label{eqn:hattheta}
\hat{\theta} = \argmin\limits_{\theta} L(\theta; \sDsan) \eqdef \argmin\limits_{\theta}  \frac{\lambda}{2} \|\theta\|_2^2 + \frac{1}{|\sDsan|} \sum_{(x,y) \in \sDsan} \ell(\theta; x, y),
\end{equation}
where $\lambda$ is a hyperparameter controlling regularization strength and $\ell$ is a convex surrogate for the 0/1-loss that $f_\theta$ incurs.
\end{enumerate}

We consider 5 different defenses that span existing approaches to data sanitization and anomaly detection:
\begin{itemize}
\item The \Ltwo{} defense removes points far from their class centroids in $L_2$ distance:
\begin{align*}
\beta_y &= \E_{\sD} [x|y]\\
s_\beta(x, y) &= \|x - \beta_y\|_2
\end{align*}

\item The \slab{} defense \citep{steinhardt2017certified} projects points onto the line between the class centroids, then removes points too far from the centroids:
\begin{align*}
\beta_y &= \E_{\sD} [x|y]\\
s_\beta(x, y) &= \left | (\beta_1 - \beta_{-1})^\top (x - \beta_y) \right |
\end{align*}
The idea is to use only the relevant dimensions in feature space to find outliers.
The \Ltwo{} defense treats all dimensions equally, whereas the \slab{} defense treats the vector between the class centroids as the only relevant dimension.

\item The \Loss{} defense discards points that are not well fit by a model trained (without any data sanitization) on the full dataset $\sD$:
\begin{align*}
\beta &= \argmin_\theta \E_{\sD} [\ell_\theta(x, y)]\\
s_\beta(x, y) &= \ell_{\beta}(x, y).
\end{align*}
It is similar to the trimmed loss defense proposed in the context of regression in \citet{jagielski2018manipulating}.
For a linear model, it is similar to the \slab{} defense, except that the relevant dimension is learned using the loss function
instead of being fixed as the direction between the class centroids.

\item The \PCA{} defense assumes that the clean data lies in some low-rank subspace,
and that poisoned data therefore will have a large component out of this subspace \citep{rubinstein2009antidote}.
Let $X$ be the data matrix, with the $i$-th row containing $x_i$, the features of the $i$-th training point. Then:
\begin{align*}
\beta &= \text{Matrix of top } k \text{ right singular vectors of } X\\
s_\beta(x, y) &= \| (I - \beta \beta^\top) x \|_2
\end{align*}
In our experiments, we choose the smallest $k$ such that the normalized Frobenius approximation error (i.e., the normalized sum of the squared singular values) is \textless 0.05.

\item The \knn{} defense removes points that are far from their $k$ nearest neighbors (e.g., \citet{frederickson2018attack}).
\begin{align*}
\beta &= \sDcp\\
s_\beta(x, y) &= \text{Distance to } k \text{-th nearest neighbor in } \beta
\end{align*}
In our experiments, we set $k = 5$.
\end{itemize}

Note that $\beta$ is sometimes a simple set of summary statistics of the dataset (e.g., in the \Ltwo{} and \slab{} defenses),
while at other times $\beta$ can be the entire dataset (e.g., in the \knn{} defense).
We will handle these two types of defenses separately, as we discuss in \refsec{attack}.

The feasible set $\sF_\beta$ encodes both the defenses and the input constraints of the dataset,
since $\sF_\beta \subseteq \sX \times \sY$ and the input domain $\sX$ only includes valid points.
Thus, the defender will eliminate all input points that do not obey the input constraints of the dataset.

%% file: attack.tex
\section{Attack Framework}
\label{sec:attack}

In this paper, we take on the role of the attacker.
Recall that we are given a set of $n$ clean training points $\sDc$ and a test set $\sDtest$,
and our goal is to come up with a set of $\epsilon n$ poisoned training points $\sDp$
such that a defender following the procedure in \refsec{defenses} will choose model parameters $\hat\theta$
that incur high test error $\sL(\hat\theta)$.
The difficulty lies in choosing poisoned points $\sDp$ that will both lead to high test error and also avoid being flagged as anomalous.

In this section, we describe our general approach to crafting attacks that can evade anomaly detectors.
As mentioned in \refsec{intro}, we can roughly group anomaly detectors into two categories:
those that are more sensitive to the data and tend to `overfit' (like the \knn{} defense),
and those that are less sensitive to the data and tend to `underfit' because they make strong parametric assumptions (like the \Ltwo{} defense).
In \refsec{concentrated}, we discuss how we can use concentrated attacks
to evade defenses in the first group.
In \refsec{optimization}, we formulate the constrained optimization problem that we use to evade defenses in the second group.
Finally, in \refsec{improved-constraints}, we introduce a randomized rounding procedure to handle problem settings where the input features are constrained to be integers.

\subsection{Concentrated attacks}\label{sec:concentrated}
To bypass anomaly detectors that are sensitive to small changes in the data,
we rely on the simple observation that poisoned data that is \emph{concentrated} on a few locations tends to appear normal to anomaly detectors.
For example:
\begin{itemize}
\item For the \knn{} defense and other similar nonparametric detectors, this is trivially true: if several poisoned points are placed very near each other,
then by definition, the distances to their nearest neighbors will be small.
\item For the \PCA{} defense, it is more likely that the low-rank representation of $\sD$ will include the poisoned points, reducing their out-of-projection components.
\item For the \Loss{} defense, if the poisoned points are concentrated in a similar location, the model will have more incentive to fit those points (because fitting one of them would imply fitting all of them, which would reduce the training loss more than fitting a single isolated point).
\end{itemize}

A potential issue for the attacker
is that being constrained to place points in concentrated groups might make the attack less effective,
in the sense of requiring more poisoned points to make the defender learn some target parameters.
For example, it might be the case that a more efficient attack would involve spreading out each poisoned point throughout the feasible set.

Fortunately for the attacker, we show that if the feasible sets for each class are convex, and if the defender is using a 2-class SVM or logistic regression model, then the above scenario will not occur.
Instead, we would only need two distinct points (one per class) to realize any attack:
\begin{theorem}[2 points suffice for 2-class SVMs and logistic regression]\label{thm:2points}
Consider a defender that learns a 2-class SVM or logistic regression model by first discarding all points outside a fixed feasible set $\sF$ and then minimizing the average (regularized) training loss.
Suppose that for each class $y = -1, +1$, the feasible set
$\sFy \eqdef \{x: (x,y) \in \sF \}$ is a convex set.
If a parameter $\hat{\theta}$ is attainable by any set of $\tilde n$ poisoned points
$\sDp = \{(\xp_1,\yp_1),\ldots,(\xp_{\tilde n}, \yp_{\tilde n})\}  \subseteq~\sF$,
then there exists a set of at most $\tilde n$ poisoned points $\tDp$ (possibly with fractional copies)
that also attains $\hat\theta$ but only contains $2$ distinct points, one from each class.

More generally, the above statement is true for any margin-based model with loss of the form $\ell(\theta; x, y) = c(-y\theta^\top x)$,
where $c: \R \to \R$ is a convex, monotone increasing, and twice-differentiable function, and the ratio of second to first derivatives $c''/c'$ is monotone non-increasing.
\end{theorem}
\begin{remark}
If the attacker concentrates all of the poisoned points in only two distinct locations, we will trivially evade the \knn{} defense.
The remaining defenses---\Ltwo{}, \slab{}, \Loss{}, and \PCA{}---all have convex feasible sets, so the conditions of the theorem hold.
One technicality is that in reality, the feasible set will not remain fixed as the theorem assumes: if an attack $\sDp$ is collapsed into
a different attack $\tDp$ with at most 2 distinct points, the feasible set $\sFb$ will also change.
However, as discussed above, it will tend to change in a way that makes the poisoned points feasible,
so if the original attack $\sDp$ was already feasible, the new attack $\tDp$ will likely also be feasible.
\end{remark}
We defer the full proof to \refapp{2points}.
As a short proof sketch, the proof consists of two parts.
We first relate the number of distinct points necessary to achieve an attack to
the geometry of the set of gradients of points within the feasible set, using the notion of \car{} numbers from convex geometry.
We then show, for the specific losses considered above (the hinge and logistic loss), that this set of feasible gradients has the necessary geometry.
Our method is general and can be extended to different loss functions and feasible sets; we provide one such extension, to a multi-class SVM setting, in \refapp{2points}.

One objection to attacks with only two distinct poisoned points is that they can be easily defeated by a defender that throws out repeated points.
However, the attacker can add a small amount of random noise to fuzz up poisoned points without sacrificing the concentrated nature of the attack.
Randomized rounding, which we use to handle datasets with integer input constraints in \refsec{improved-constraints},
is a version of this procedure.

\subsection{Constrained optimization}\label{sec:optimization}
Anomaly detectors that are more robust to small changes in the training data, such as those that rely on simple sufficient statistics of the data,
tend to be less vulnerable to concentrated attacks.
In our setting, the \Ltwo{} and \slab{} defenses in particular are not fooled by concentrated attacks:
the class centroid, which is used in both defenses,
cannot be moved too much by an $\epsilon$ fraction of poisoned points if $\epsilon$ is small and the clean training data $\sDc$ is well-clustered,
since poisoned points that are too far away would be filtered out by the \Ltwo{} defense.

To handle these defenses, we formulate the attacker's goal as a constrained optimization problem:
\begin{align}
\label{eqn:general-opt}
\underset{\sDp}{\text{maximize}} \quad
& \Lerr(\hat\theta; \sDtest) \\
\text{s.t.} \quad
& |\sDp| = \epsilon |\sDc|, &\text{($\epsilon$ fraction of poisoned points)}\nonumber\\
\text{where} \quad
& \beta = B(\sDcp) &\text{($\sFb$ is fit on clean and poisoned data)}\nonumber\\
& \hat{\theta} \eqdef \argmin_{\theta} L \left( \theta; (\sDcp)\cap\sF_\beta \right). &\text{(defender trains on remaining data)}\nonumber
\end{align}
The first constraint corresponds to the attacker only being able to add in an $\epsilon$ fraction of poisoned points;
the second constraint corresponds to the defender fitting the anomaly detector on the entire training set $\sDcp$;
and the final equality corresponds to the defender learning model parameters $\hat\theta$ that minimize training loss.

To make this problem more tractable, we make three approximations:
\begin{enumerate}
  \item We assume that the defender does not discard any clean points. Thus, if all poisoned points are constrained to lie within the feasible set $\sF_\beta$
  and therefore evade sanitization, then the defender trains on the entirety of $\sD = \sDc \cup \sDp$.%
  \footnote{This favors the defender,
  since we do not consider the case in which a savvy attacker might place poisoned points in such a way as to
  cause the defender to throw out particularly good points in $\sDc$
  and therefore learn a bad model.}
  \item We break the dependence of the feasible set $\sF_\beta$ on
  the poisoned data $\sDp$ by fixing the anomaly detector $\beta = B(\sDc)$ on the clean data. This is a reasonable approximation for the \Ltwo{} and \slab{} defenses, as their feasible sets depend only on the class centroids, which are very stable with respect to small amounts of poisoned data $\sDp$.
  \item Finally, as is standard, we replace the 0-1 test error $\Lerr(\hat\theta; \sDtest)$ with its convex surrogate $L(\hat\theta; \sDtest) \eqdef \E_{\sDtest} \ell(\hat\theta; x, y)$, which is continuous and
  easier to optimize.
\end{enumerate}
These approximations let us rewrite the attacker's goal as the following optimization problem:
\begin{align}
\label{eqn:attacker-opt}
\underset{\sDp}{\text{maximize}} \quad
& L(\hat\theta; \sDtest) \\
\text{s.t.} \quad
& |\sDp| = \epsilon |\sDc|\nonumber\\
& \sDp \subseteq \sF_\beta\nonumber\\
\text{where} \quad
& \hat{\theta} \eqdef \argmin_{\theta} L(\theta; \sDcp) \nonumber\\
& \beta = B(\sDc). \nonumber
\end{align}
This constrained optimization formulation has been used in prior work (e.g., \citet{steinhardt2017certified}),
and despite the approximations, it is still a bilevel problem that is non-convex and intractable to solve exactly \citep{bard1991some, bard1999}.
Our contribution in this regard is developing more effective and computationally efficient ways of solving it,
which will be the focus of \refsec{specific_attacks}.

Instead of fixing the feasible set based on only the clean data, $\beta = B(\sDc)$, we can also adopt an iterative optimization approach,
where we alternate between optimizing over $\sDp$ for a fixed $\beta$, and then updating $\beta$ to reflect the new $\sDp$ (Algorithm~\ref{alg:constrained-opt}).
This iterative optimization procedure is guaranteed to make progress so long as the poisoned points $\sDp$
remain valid even after re-fitting $\beta$.
In \refsec{expt_iterative}, we show that this slightly improves the attacks obtained, though it is not necessary to obtain effective attacks.

\begin{algorithm}
\caption{Pseudocode for attack via iterative constrained optimization.}
\label{alg:constrained-opt}
\begin{algorithmic}
\STATE Input: clean data $\sDc$, poisoned fraction $\epsilon$.
\STATE Initialize anomaly detector on clean data: $\beta \gets B(\sDc)$
\REPEAT
   \STATE
  $\sDp \gets \underset{\substack{  |\sDp| = \epsilon |\sDc| \\ \sDp \subseteq \sF_\beta }} \argmax   L(\hat\theta; \sDtest)$
  \hspace{8mm} (Solve \refeqn{attacker-opt} with fixed $\beta$; see Sections \ref{sec:improved-inf}-\ref{sec:improved-mm})
  \vspace{2mm}
  \STATE $\beta \gets B(\sDcp)$
  \hspace{28.1mm} (Update anomaly detector $\beta$ with new $\sDp$)
\UNTIL{$\sDp$ converges}
\STATE Output $\sDp$.
\end{algorithmic}
\end{algorithm}

\subsection{Handling integer input constraints with randomized rounding}
\label{sec:improved-constraints}
Each of the three data poisoning attacks that we will develop use some form of gradient descent on the poisoned points to solve the attacker's optimization problem \refeqn{attacker-opt}.
However, gradient descent cannot be directly applied in settings where the input features are constrained to be non-negative integers (e.g., with bag-of-word models in natural language tasks).

To handle this, \citet{steinhardt2017certified} relaxed the integrality constraint to allow non-negative real-valued points, and then performed randomized rounding on these real-valued points to obtain integer-valued points:
\begin{enumerate}
\item Solve the optimization problem \refeqn{attacker-opt} while allowing all poisoned points $(x, y) \in \sDp$
to have real-valued $x$.
\item Then, for each $(x, y) \in \sDp$, we construct a rounded $\hat x$ as follows:
for each coordinate $i$, $\hat{x}_i = \lceil x_i \rceil$ with probability
$x_i - \lfloor x_i \rfloor$, and $\hat{x}_i = \lfloor x_i \rfloor$ otherwise. This procedure preserves the mean: $\E[\hat{x}] = x$.
\end{enumerate}
However, this approach can produce poisoned points that get detected by data sanitization defenses. We adopt their approach but introduce two techniques to avoid detection:

\paragraph{Repeated points.}
In high dimensions, randomized rounding can result in poisoned points that are far away from each other.
This can make the poisoned points vulnerable to being filtered out by the defender, as they would no longer be concentrated in a few distinct locations (\refsec{concentrated}).
To address this, we adopt a heuristic of repeating poisoned points $r$ times after rounding
(keeping the overall fraction of poisoned data, $\epsilon$, the same).
This means that we find $\epsilon n / r$ poisoned points $\{x_1, x_2, \ldots, x_{\epsilon n / r}\}$,
do randomized rounding to get $\{\hat x_1, \hat x_2, \ldots, \hat x_{\epsilon n / r}\}$, and then form
the multiset $\sDp$ by taking the union of $r$ copies of this set.
In practice, we find that setting $r=2$ or 3 works well.
This heuristic better concentrates the poisoned points
while still preserving their expected mean $\E[\hat x]$.

\paragraph{Linear programming (LP) relaxation for the \Ltwo{} defense.}
Randomized rounding violates the $L_2$ constraint used in the \Ltwo{} defense.
Recall that in the \Ltwo{} defense, we wish to play points
$(x,y)$ such that $\|x - \mu_y\|_2 \leq \tau_y$, where $\mu_y$
is the mean of class $y$ and $\tau_y$ is some threshold (\refsec{defenses}).
The issue is Jensen's inequality: since the $L_2$ norm is convex,
$\E[\|\hat{x} - \mu_y\|_2] \geq \|\E[\hat{x}] - \mu_y\|_2 = \|x - \mu_y\|_2$.
This means that even if we control the norm of the continuous $x$
by having $\|x - \mu_y\|_2 \leq \tau_y$, we could still have the randomly-rounded $\hat x$ violate this constraint on expectation: $\E[\|\hat{x} - \mu_y\|_2] > \tau_y$.

\citet{steinhardt2017certified} deal with this problem by setting $\tau_y$ conservatively,
so that $\hat{x}$ might avoid detection even if $\|\hat{x} - \mu_y\|_2 > \|x - \mu_y\|_2$.
However, the conservative threshold reduces the attacker's options, resulting in a less effective attack.
Instead, our approach is to control the expected squared norm $\E[\|\hat{x} - \mu_y\|_2^2]$:
if $\E[\|\hat{x} - \mu_y\|_2^2] < \tau_y^2$, then by Jensen's inequality, $\E[\|\hat{x} - \mu_y\|_2] < \tau_y$.
To compute the expected squared norm, we first write
\begin{align}
\E[\|\hat{x} - \mu_y\|_2^2]
 &= \E[\|\hat{x}\|_2^2] - 2\langle x, \mu_y \rangle + \|\mu_y\|_2^2.
\end{align}

\begin{figure}[h]
\begin{center}
\includegraphics[width=0.45\textwidth]{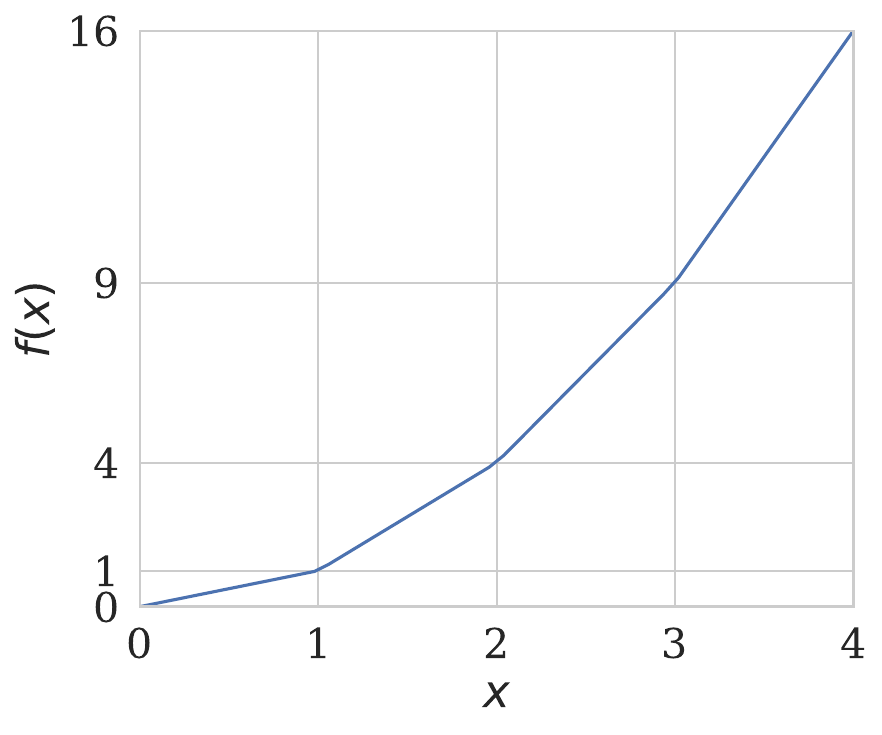}
\vspace{-3mm}
\caption{Plot of $\E[\|\hat{x}\|_2^2] = f(x)$ against $x$ for scalar $x$.}
\label{fig:lp-function}
\end{center}
\end{figure}

Note that $\E[\|\hat{x}\|_2^2]$ can in general be substantially larger than $\|x\|_2^2$ due
to the variance from rounding.
We can compute $\E[\|\hat{x}\|_2^2]$ explicitly as
\begin{align}
\E[\|\hat{x}\|_2^2] &= \sum_{i=1}^d x_i(\lceil x_i \rceil + \lfloor x_i \rfloor) - \lceil x_i \rceil \lfloor x_i \rfloor.
\end{align}
While the function $f(x) \eqdef x(\ceil{x} + \floor{x}) - \ceil{x} \floor{x}$ looks complicated, intuitively, we expect $f(x)$ to be close to $x^2$. Indeed,
as \reffig{lp-function} shows, it is a piecewise-linear function where the $k$-th piece linearly interpolates
between $k^2$ and $(k+1)^2$,
and we can write it as the maximum over a set of linear equations:
\begin{equation}
\label{eq:f-lp}
f(x) = \max_{k=0}^{\infty} (2k+1)x - k(k+1).
\end{equation}

Thus, when solving the attacker optimization (equation \refeqn{attacker-opt}) for datasets with non-negative integer constraints,
we replace the standard \Ltwo{} feasible set $\sFb = \{(x,y) : \|x-\mu_y\|_2 \leq \tau_2 \text{ and } x \in \bR_{\geq 0}\}$
with the modified constraint set
\begin{equation}
\sF_{\text{LP}} = \{(x,y) : \E[\|\hat{x}-\mu_y\|_2^2] \leq \tau_y^2 \text{ and } x \in \bR_{\geq 0}\}.
\end{equation}
If we approximate the infinite maximum in \eqref{eq:f-lp} by its first
$M$ terms, then the corresponding approximation of $\sF_{\text{LP}}$ can be represented
via a linear program. In our experiments,
we choose $M$ adaptively for each coordinate $i$ to be equal to the largest value that
$x_i$ attains across the dataset.
This formulation allows us to express the $L_2$ norm constraint $\E[\|\hat{x} - \mu_y \|_2^2] \leq \tau_y^2$ as a set of linear constraints on the continuous $x$,
allowing us to control the expected $L_2$ norm of the poisoned points after rounding.

Randomized rounding has some negative impact: it makes attacks less concentrated, as discussed above,
and can also result in a few unlucky poisoned points getting filtered by other defenses (e.g., by the \Loss{} defense if the rounding happens to increase the loss on the point).
Another advantage of the above LP relaxation is that in practice, the linear constraints tend to lead to nearly-integer $x$,
which further reduces the negative impact of having to do randomized rounding.

%% file: specific_attacks.tex
\section{Specific Attacks}
\label{sec:specific_attacks}

In this section, we introduce three different methods for efficiently solving the attacker's optimization problem (Equation \refeqn{attacker-opt}) and generating an attack.
As discussed in \refsec{attack}, these methods all generate concentrated attacks within our constrained optimization framework and use the randomized rounding procedure when necessary.
We start with the \inf{} attack in \refsec{improved-inf}, which is direct but computationally slower, and then introduce the \kkt{} and \mm{} attacks in \refsecs{kkt}{improved-mm}, which both use the idea of decoy parameters to speed up the attack.

\subsection{The \inf{} attack}\label{sec:improved-inf}

Recall that solving Equation \refeqn{attacker-opt} involves finding poisoned data $\sDp$ that maximizes the defender's test loss $L(\hat\theta; \sDtest)$ for a fixed feasible set $\sF_\beta$.
The \inf{} attack tackles this problem via projected gradient ascent.

At a high level, we can find a local maximum of Equation \refeqn{attacker-opt} by iteratively taking gradient steps on the features of each poisoned point in $\sDp$,
projecting each point onto the feasible set $\sF_\beta$ after each iteration.
This type of gradient-based data poisoning attack was first studied in the context of SVMs by \citet{biggio2012poisoning},
and has subsequently been extended to
linear and logistic regression \citep{mei2015teaching},
topic modeling \citep{mei2015security},
collaborative filtering \citep{li2016data},
and neural networks \citep{koh2017understanding, yang2017generative, munoz2017towards}.
We call this projected gradient ascent method the influence attack after
\citet{koh2017understanding}, who use influence functions to compute this gradient.
Our method builds upon previous work by incorporating the techniques mentioned in \refsec{attack}---concentrating the attack and using randomized rounding with the LP relaxation.

\subsubsection{The basic influence attack}
First, we review the basic influence attack, borrowing from the presentation in \citet{koh2017understanding}.
Our goal is to perform gradient ascent on each poisoned point $(\xp, \yp) \in \sDp$ to maximize the test loss $L(\hat\theta; \sDtest)$.
The \infbasic{} algorithm (Algorithm~\ref{alg:influence}) iteratively computes the gradient $\frac{\partial L}{\partial \xp}$ for
each poisoned point $(\xt, \yt) \in \sDp$, moves each point in the direction of its gradient, and then projects each point back onto the feasible set $\sF_\beta$.
We only optimize over the input features $\xp$; since the labels $\yp$ are discrete, we cannot compute gradients on them.
Instead, we fix the labels $\yp$ at the beginning of the algorithm, grid searching over the fraction of positive versus negative labels.
We provide more implementation details in \refapp{app_inf_details}.

\paragraph{Computing the gradient.}
The difficulty in computing the gradient of the test loss $L(\hat\theta; \sDtest)$ w.r.t. each $\xp$ in $\sDp$ is that
$L$ depends on $\xp$ only through the model parameters $\hat\theta$,
which is a complicated function of $\sDp$.
The influence attack uses a closed-form estimate of $\frac{\partial \hat{\theta}}{\partial \xp}$, which measures how much the model parameters $\hat\theta$ change with a small change to $\xp$.
The desired derivative $\frac{\partial L}{\partial \xp}$ can then be computed via the chain rule:
$\frac{\partial L}{\partial \xp} = \frac{\partial L}{\partial \hat{\theta}} \frac{\partial \hat{\theta}}{\partial \xp}$.

The quantity $\frac{\partial L}{\partial \hat{\theta}}$ is the average gradient of the test loss, which we denote as $\gtest$ for convenience,
and it can be computed straightforwardly as
\begin{equation}
\label{eqn:g-def}
\gtest \eqdef \frac{\partial L}{\partial \hat{\theta}} = \frac{1}{|\sDtest|} \sum_{(x,y) \in \sDtest} \nabla \ell(\hat\theta; x,y).
\end{equation}
Calculating $\frac{\partial \hat{\theta}}{\partial \xp}$ is more involved, but a standard result (see, e.g., Section 2.2 of \citet{koh2017understanding})
gives the expression
\begin{equation}
\label{eqn:influence-params}
\frac{\partial \hat{\theta}}{\partial \xp} = -H_{\hat{\theta}}^{-1} \frac{\partial^2 \ell(\hat{\theta}; \xp, \yp)}{\partial \hat{\theta} \, \partial \xp},
\end{equation}
where $H_{\hat{\theta}}$ is the Hessian of the training loss at $\hat{\theta}$:
\begin{equation}
\label{eqn:H-def}
H_{\hat{\theta}} \eqdef \lambda I + \frac{1}{|\sDc \cup \sDp|} \sum_{(x,y) \in \sDc \cup \sDp} \frac{\partial^2 \ell(\hat{\theta}; x,y)}{\partial \hat{\theta}^2}.
\end{equation}
Combining equations \refeqn{g-def} to \refeqn{H-def}, the gradient of the test loss w.r.t. an attack point $\xp$ is
\begin{align*}
\frac{\partial L(\hat{\theta})}{\partial \xp} = -\gtest ^{\top} H_{\hat{\theta}}^{-1} \frac{\partial^2 \ell(\hat{\theta}; \xp, \yp)}{\partial \hat{\theta} \, \partial \xp}.
\end{align*}

\paragraph{Projecting onto the feasible set.}
To prevent the poisoned points from being sanitized, we project each poisoned point $(\xt, \yt) \in \sDp$ onto the feasible set $\sF_\beta$.
The projection is well-defined for the \Ltwo{} and \slab{} defenses, as their feasible sets are convex, and finding the projection is a convex optimization problem that can be solved efficiently by a general-purpose convex solver.
When the $\xp$'s are constrained to take on integer values, the \infbasic{} attack uses the vanilla randomized rounding procedure described in \refsec{improved-constraints} (without the repeated points heuristic or linear programming relaxation).

\begin{algorithm}
\caption{The \infbasic{} attack.}
\label{alg:influence}
\begin{algorithmic}
\STATE Input: clean data set $\sDc$, poisoning fraction $\epsilon$, step size $\eta$, feasible set $\sF_\beta$.
\STATE Initialize poisoned data set $\sDp \gets \{(\xp_1, \yp_1), \ldots, (\xp_{\epsilon n}, \yp_{\epsilon n})\}$.
\FOR{$t = 1, 2, \ldots$}
  \STATE Compute model parameters $\hat\theta \gets \argmin_{\theta} L \left( \theta; (\sDcp)\cap\sF_\beta \right)$.
  \STATE Pre-compute $\gtest^{\top} H_{\hat{\theta}}^{-1}$ as in \eqref{eqn:g-def} and \eqref{eqn:H-def}.
  \FOR{$i = 1, \ldots, \epsilon n$}
    \STATE Set $\xp_i^0 \gets \xp_i - \eta \gtest^{\top}H_{\hat\theta}^{-1}\frac{\partial^2 \ell(\hat{\theta}; \xp_i, \yp_i)}{\partial \hat{\theta} \, \partial \xp_i}$.
    \hspace{6mm} (gradient update)
    \STATE Set $\xp_i \gets \argmin_{x \in \sF_\beta} \|x - \xp_i^0\|_2$.
    \hspace{15.5mm} (projection)
  \ENDFOR
\ENDFOR
\STATE Output $\sDp$.
\end{algorithmic}
\end{algorithm}

\subsubsection{Improvements to the basic algorithm}
We introduce the \inf{} attack, which improves upon the \infbasic{} attack in two ways. First, it incorporates randomized rounding with the LP relaxation, as described in \refsec{improved-constraints}.
Second, it concentrates the attack (\refsec{concentrated}), which makes the attack stronger and more computationally efficient.
The \infbasic{} attack optimizes over each of the $\epsilon n$ poisoned points separately, which is slow and results in poisoned points that are often quite far from each other (because of differences in initialization),
leaving them vulnerable to being detected as anomalies.
As \refthm{2points} shows,
we can modify the algorithm to instead only consider copies of two distinct points $(\xpp, 1)$ and $(\xpn, {-1})$, one from each class, without any loss in potential attack effectiveness.
This is faster, as at each iteration, we only need to compute the gradients and do the projection twice (vs. $\epsilon n$ times).
Moreover, the resulting attack is by construction concentrated on only two distinct points, helping it evade detection.

However, even after these improvements, the \inf{} attack is slow especially in high dimensions: each iteration of gradient descent requires computing an expensive
inverse Hessian-vector product \refeqn{influence-params} and a projection onto the feasible set.
Moreover, the \inf{} attack relies on local optimization and can sometimes get stuck in poor local minima, even when the underlying model loss is convex.
To mitigate these shortcomings, we propose the \kkt{} attack, which we discuss next.

\subsection{The \kkt{} attack}\label{sec:kkt}
The \kkt{} attack is based on the observation that the attacker's optimization problem \refeqn{attacker-opt} is difficult to solve because the optimization variable $\sDp$ only affects the objective (test loss $L(\hat\theta; \sDtest)$) through the model parameters $\hat\theta$,
which are themselves a complicated function of $\sDp$.
In general, we do not know what $\hat\theta$ would lead to an attack that is both effective and realizable;
but if we did know which $\hat\theta$ we were after,
then the attacker's optimization problem simplifies to finding $\sDp$ such that $\hat\theta = \argmin_{\theta} L(\theta; \sDcp)$.
As we will show in this section, this simplified problem can be solved much more efficiently than the original bilevel problem.

The \kkt{} attack has two parts:
\begin{enumerate}
  \item Using fast heuristics to find \emph{decoy parameters} $\thetadecoy$ that we want the defender to learn, and then
  \item Finding poisoned data $\sDp$ that tricks the defender into learning those decoy parameters $\thetadecoy$.
\end{enumerate}
The name of this attack comes from the use of the Karush-Kuhn-Tucker (KKT) first-order
necessary conditions for optimality in the second step.

\subsubsection{Finding decoy parameters $\thetadecoy$}
\label{sec:kkt-choosing-decoys}
Good decoy parameters, from the perspective of the attacker, should have high test error while still being achievable by some poisoned data $\sDp$. Decoy parameters that have a high loss on the clean data $\sDc$ are unlikely to be achievable by an $\epsilon$ fraction of poisoned data $\sDp$,
since it is likely that there exist other parameters that would have a lower training loss on the combined data $\sDcp$ and would therefore be learned instead by the defender.

\begin{algorithm}
\caption{Finding decoy parameters $\thetadecoy$}
\label{alg:decoy}
\begin{algorithmic}
\STATE Input: clean data set $\sDc$, loss threshold $\gamma$, number of repeats $r$.
  \STATE $\thetac \gets \argmin_\theta L(\theta; \sDc)$
  \STATE $\sDflip \gets r \text{ copies of } \{(x, y) \in \sDtest: \ell(\thetac; x, y) \geq \gamma\}$
  \STATE $\sDdecoy \gets \sDc \cup \sDflip$
  \STATE $\thetadecoy \gets \argmin_\theta L(\theta; \sDdecoy)$
\STATE Output $\sDdecoy$.
\end{algorithmic}
\end{algorithm}

Our heuristic is to augment the clean data $\sDc$ with a dataset $\sDflip$ comprising label-flipped examples from $\sDtest$,
and then find the parameters $\thetadecoy$ that minimize the training loss on this modified training set $\sDdecoy = \sDc \cup \sDflip$ (\refalg{decoy}).
The idea is that since the decoy parameters $\thetadecoy$ were trained on $\sDdecoy$, which incorporates flipped points from $\sDtest$,
it might achieve high test loss. At the same time, the following informal argument suggests that on the clean data $\sDc$, the decoy parameters $\thetadecoy$ are likely to be not much worse than the optimal parameters for the clean data, $\thetac$.
By construction,
\begin{align*}
|\sDc|\cdot L(\thetadecoy; \sDc) + |\sDflip|\cdot L(\thetadecoy; \sDflip) &= |\sDdecoy|\cdot L(\thetadecoy; \sDdecoy)\\
&\leq  |\sDdecoy|\cdot L(\thetac; \sDdecoy)\\
&= |\sDc|\cdot L(\thetac; \sDc) + |\sDflip|\cdot L(\thetac; \sDflip).
\end{align*}
Rearranging terms, this implies that
\begin{align*}
L(\thetadecoy; \sDc) &\leq L(\thetac; \sDc) + \frac{|\sDflip|}{|\sDc|} \cdot \left(L(\thetac; \sDflip) - L(\thetadecoy; \sDflip)\right)\\
&\leq L(\thetac; \sDc) + \frac{|\sDflip|}{|\sDc|} \cdot L(\thetac; \sDflip),
\end{align*}
where the last inequality comes from the non-negativity of the loss $L$.
The second term on the right-hand side, $L(\thetac; \sDflip)$, is likely to be small:
$\sDflip$ comprises of points that originally had a high loss under $\thetac$ before their labels were flipped
(which implies that their label-flipped versions are likely to have a lower loss), and we can choose $|\sDflip|$ to be small compared to $|\sDc|$.
Thus, the average loss $L(\thetadecoy; \sDc)$ of the decoy parameters $\thetadecoy$ on the clean data $\sDc$ is not likely to be too much higher than $L(\thetac; \sDc)$, which is the best possible average loss on $\sDc$ within the model family.

By varying the loss threshold $\gamma$ and number of repeats $r$ used in \refalg{decoy}, we obtain different candidates for $\thetadecoy$.
As we will discuss next, finding an attack $\sDp$ for each candidate $\thetadecoy$ is fast,
so we simply generate a set of candidate decoy parameters and try all of them, picking the $\thetadecoy$ that achieves the highest test loss.

\subsubsection{Attacking with known $\thetadecoy$}
For given decoy parameters $\thetadecoy$, the next step for the attacker is to find poisoned data $\sDp$ such that $\sDp$ evades data sanitization and $\thetadecoy$ minimizes the overall training loss over
both $\sDp$ and the clean data $\sDc$. We can formulate this task as the following optimization problem:
\begin{align}
\label{eqn:kkt-with-argmin}
\text{find} \quad
& \sDp \\
\text{s.t.} \quad
& |\sDp| = \epsilon |\sDc|\nonumber\\
& \sDp \subseteq \sF_\beta \nonumber\\
& \thetadecoy = \argmin_{\theta} L(\theta; \sDcp). \nonumber
\end{align}
Since $\thetadecoy$ is pre-specified, we can rewrite the inner optimization as a simple equality.
Specifically, if the loss $\ell$ is strictly convex and differentiable in $\theta$,
we can rewrite the condition
\begin{align*}
\thetadecoy &= \argmin_{\theta} L(\theta; \sDcp)\\
&= \argmin_{\theta} \sum_{(x,y) \in \sDc} \ell(\theta; x, y) + \sum_{(\xp,\yp) \in \sDp} \ell(\theta; \xp, \yp)
\end{align*}
as the equivalent KKT optimality condition
\begin{align}
\label{eqn:kkt-condition}
\sum_{(x,y) \in \sDc} \nabla_\theta \ell(\thetadecoy; x, y) + \sum_{(\xp,\yp) \in \sDp} \nabla_\theta \ell(\thetadecoy; \xp, \yp) = 0.
\end{align}
If the loss $\ell$ is not differentiable, e.g., the hinge loss, we can replace this with a similar subgradient condition.

Since the first term in \refeqn{kkt-condition} is fixed and does not depend on the optimization variable $\sDp$,
we can treat it as a constant:
$$\gDc \eqdef \inv{|\sDc|} \sum_{(x,y) \in \sDc} \nabla_\theta \ell(\thetadecoy; x, y).$$
Rewriting \refeqn{kkt-condition} as $\gDc + \inv{|\sDc|} \sum_{(\xp,\yp) \in \sDp} \nabla_\theta \ell(\thetadecoy; \xp, \yp) = 0$
and substituting it into \refeqn{kkt-with-argmin} gives us
\begin{align}
\label{eqn:kkt-without-argmin}
\text{find} \quad
& \sDp \\
\text{s.t.} \quad
& |\sDp| = \epsilon |\sDc|\nonumber\\
& \sDp \subseteq \sF_\beta \nonumber\\
&\gDc + \inv{|\sDc|} \sum_{(\xp,\yp) \in \sDp} \nabla_\theta \ell(\thetadecoy; \xp, \yp) = 0. \nonumber
\end{align}
If this optimization problem \refeqn{kkt-without-argmin} has a solution, we can find it by solving the equivalent
norm-minimization problem
\begin{align}
\label{eqn:kkt-opt}
\underset{\sDp}{\text{minimize}} \quad
& \Big\| \gDc + \inv{|\sDc|} \sum_{(\xp,\yp) \in \sDp} \nabla_\theta \ell(\thetadecoy; \xp, \yp) \Big\|^2_2 \\
\text{s.t.} \quad
& |\sDp| = \epsilon |\sDc|\nonumber\\
& \sDp \subseteq \sF_\beta, \nonumber
\end{align}
which moves the KKT constraint into the objective,
relying on the fact that the norm of a vector is minimized when the vector is 0.

Next, we make use of \refthm{2points}, which shows that for the binary classification
problems we consider, we can concentrate our attacks by placing all of the poisoned
points at two distinct locations $\xpp$ and $\xpn$ without any loss in attack effectiveness.
If we let $\epp \cdot n$ and $\epn \cdot n$ be the number of positive and negative poisoned points added, respectively,
we can write \refeqn{kkt-opt} as
\begin{align}
\label{eqn:kkt-opt-twopoints}
\underset{\xpp, \xpn, \epp, \epn}{\text{minimize}} \quad
& \big\| \gDc + \epp \nabla_\theta \ell(\thetadecoy; \xpp, 1) +  \epn \nabla_\theta \ell(\thetadecoy; \xpn, -1) \big\|^2_2 \\
\text{s.t.} \quad
& \epp + \epn = \epsilon \nonumber\\
& (\xpp, 1), (\xpn, -1) \in \sF_\beta. \nonumber
\end{align}
For general losses, this optimization problem is non-convex but can be solved by local methods like gradient descent.
In our experiments, the defender uses the hinge loss $\ell(\theta; x, y) = \max(0, 1 - y \theta^{\top}x)$ and applies $\ell_2$ regularization with regularization parameter $\lambda$ (\refeqn{hattheta}, \refsec{defenses}). This setting allows us to further rewrite \refeqn{kkt-opt-twopoints} as the following:
\begin{align}
\label{eqn:kkt-opt-svm}
\underset{\xpp, \xpn, \epp, \epn}{\text{minimize}} \quad
& \big\| \gDc - \epp \xpp +  \epn \xpn + \lambda \thetadecoy \big\|^2_2 \\
\text{s.t.} \quad
& 1 - \theta^\top \xpp \geq 0\nonumber\\
& 1 + \theta^\top \xpp \geq 0\nonumber\\
& \epp + \epn = \epsilon \nonumber\\
& (\xpp, 1), (\xpn, -1) \in \sF_\beta,\nonumber
\end{align}
where the first two constraints ensure that $(\xpp, 1)$ and $(\xpn, -1)$ are both support vectors.
This problem is convex when the feasible set $\sFb$ is convex and $\epp$ and $\epn$ are fixed;
since $\sFb$ is convex in our setting, we can grid search over $\epp$ and $\epn$
and call a generic solver for the resulting convex problem.
Pseudocode is given in Algorithm~\ref{alg:kkt}.

\begin{algorithm}
\caption{KKT attack with grid search.}
\label{alg:kkt}
\begin{algorithmic}
\STATE Input: clean data set $\sDc$, feasible set $\sF_\beta$, poisoning fraction $\epsilon$, decoy parameter $\thetadecoy$, grid search size $T$.
\FOR{$t = 0, \ldots, T$}
  \STATE Set $\epp \gets t \epsilon /T$, $\epn \gets \epsilon - \epp$.
  \STATE Obtain $\hat\theta, \xpp$, and $\xpn$ by solving \refeqn{kkt-opt-svm} with fixed values of $\epp, \epn$.
  \STATE Evaluate test error $\sL(\hat\theta)$.
\ENDFOR
\STATE Pick $\xpp, \xpn, \epp, \epn$ corresponding to highest test error $\sL(\hat\theta)$ found in grid search.
\STATE Output $\sDp = \{\epp |\sDc|$ copies of $\xpp$ and $\epn |\sDc|$ copies of $\xpn\}$.
\end{algorithmic}
\end{algorithm}

\paragraph{Evading the \Loss{} defense.}
Defenses like the \Loss{} defense have feasible sets that depend on the model parameters $\hat\theta$ that the defender learns, which in turn depend on the poisoned points.
This dependence makes it difficult for the attacker to explicitly constrain their poisoned points to lie within such feasible sets.
In the \inf{} attack, we relied on concentrated attacks (\refsec{concentrated}) to evade the \Loss{} defense.
This approach is empirically effective, but it relies to some extent on luck, as
the attacker cannot guarantee that its poisoned points will have low loss.

One advantage of decoy parameters is that they give attackers a computationally tractable handle on parameter-dependent defenses like the \Loss{} defense.
With decoy parameters, the attacker can approximately specify the feasible set $\sF_\beta$ independently of the learned parameters $\hat\theta$,
since we know that if the attack works, the learned parameters $\hat\theta$ should be close to the decoy parameters $\thetadecoy$.
For example, we can handle the \Loss{} defense by adding the constraint $\ell(\thetadecoy; \xp, \yp) < \tau_{\yp}$ to the feasible set $\sF_\beta$.

\subsection{Improved Min-Max Attack}
\label{sec:improved-mm}
Our third and final attack is the \mm{} attack, which improves on what we call the \mmbasic{} attack from prior work \citep{steinhardt2017certified}.
The \mm{} attack relies on the same decoy parameters introduced in \refsec{kkt},
but unlike the \inf{} and \kkt{} attacks, it naturally handles multi-class problems without a grid search, and it does not require convexity of the feasible set.
Its drawback is assuming that the clean data $\sDc$ and the test data $\sDtest$ are drawn
from the same distribution (i.e., that the attacker is performing an \emph{indiscriminate} attack; see \refsecs{problem}{related}).

\subsubsection{The \mmbasic{} attack}
We start by reviewing the \mmbasic{} attack, as it was introduced in \citet{steinhardt2017certified}.
Recall that the attacker's goal is to find poisoned points $\sDp$ that maximize the test loss $L(\hat\theta; \sDtest)$ that the defender incurs, where the parameters $\hat\theta$ are chosen to minimize the training loss $L(\hat\theta; \sDcp)$ (equation \refeqn{attacker-opt}).
As we discussed in Sections \ref{sec:optimization}, \ref{sec:improved-inf}, and \ref{sec:kkt}, the bilevel nature of this optimization problem---maximizing the loss involves an inner minimization to find the parameters $\hat\theta$---makes it difficult to solve.

The key insight in \citet{steinhardt2017certified} was that we can make this problem tractable by replacing the test loss
$L(\hat\theta; \sDtest)$
with the training loss $L(\hat\theta; \sDcp)$.
This substitution changes the bilevel problem into a \emph{saddle-point problem}---i.e., one that can be expressed in the form $\min_u \max_v f(u, v)$---
that can be solved efficiently via gradient descent.

To do so, we first approximate the average test loss with the average clean training loss:
\begin{align}\label{eqn:test-approx}
L(\theta; \sDtest) \approx L(\theta; \sDc).
\end{align}
This approximation only works in the setting where the test data $\sDtest$ is drawn from the same distribution as the (clean) training data $\sDc$,
and relies on the training set being sufficiently big and the model being appropriately regularized, such that test loss is similar to training loss.
Next, we make use of the non-negativity of the loss to upper bound the average clean training loss with the average combined loss on the clean and poisoned data:
\begin{align}\label{eqn:loss-bound}
L(\theta; \sDc) \leq L(\theta; \sDc) + \epsilon L(\theta; \sDp) = (1 + \epsilon) L(\theta; \sDcp),
\end{align}
where, as usual, $\epsilon$ is the relative ratio of poisoned points $\epsilon = \frac{|\sDp|}{|\sDc|}$.

By combining \refeqn{test-approx} and \refeqn{loss-bound}, we can approximately upper-bound the average test loss
$L(\theta; \sDtest)$ by $(1 + \epsilon)$ times the average loss on the combined training data $L(\theta; \sDcp)$.
Instead of directly optimizing for $L(\theta; \sDtest)$ as the attacker (equation \refeqn{attacker-opt}), we can therefore optimize for $L(\theta; \sDcp)$,
which gives us
\begin{align*}
\underset{\sDp \subseteq \sF_\beta}{\text{maximize}} \quad
& L(\theta; \sDcp) \\
\text{where} \quad
& \hat{\theta} \eqdef \argmin_{\theta} L(\theta; \sDcp). \nonumber
\end{align*}
The advantage of this formulation is that the outer maximization and inner minimization are over the same function $L(\theta; \sDcp)$, which lets us rewrite it as the saddle-point problem
\begin{equation}
\label{eqn:saddle}
\underset{\sDp \subseteq \sF_\beta}{\max}
\underset{\theta}{\min} \
L(\hat\theta; \sDcp).
\end{equation}

When the loss $\ell$ is convex, we can solve \refeqn{saddle}
by swapping min and max and solving the resulting problem $\min_\theta \max_{\sDp \subseteq \sF_\beta} L(\theta; \sDcp)$, which expands out to
\begin{equation}
\label{eq:mm}
\min_{\theta} \Big[ \frac{\lambda (1 + \epsilon)}{2} \|\theta\|_2^2 + \frac{1}{|\sDc|} \sum_{(x,y) \in \sDc} \ell(\theta; x,y) + \epsilon \max_{(\xp,\yp) \in \sF_\beta} \ell(\theta; \xp, \yp) \Big].
\end{equation}
This problem is convex when the loss $\ell$ is convex, and we can solve it via subgradient descent by iteratively
finding $(\xp, \yp) \in \sF_\beta$ that maximizes $\ell(\theta; \xp, \yp)$,
then taking the subgradient of the outer expression w.r.t. $(\xp, \yp)$.

To solve the inner problem of finding $(\xp, \yp) \in \sFb$ that maximizes $\ell(\theta; \xp, \yp)$,
we note that if the model is margin-based, i.e., $\ell(\theta; \xp, \yp) = c(-y\theta^\top x)$ for some monotone increasing function $c$ (which is the case for SVMs and logistic regression), then maximizing $\ell(\theta; \xp, \yp)$ is equivalent to minimizing the margin $y\theta^\top x$.
For a fixed $\yp$, we can solve the convex problem
\begin{align*}
\underset{\xp}{\text{minimize}} \quad
& \yp\theta^\top \xp\\
\text{s.t.} \quad
& (\xp, \yp) \in \sFb.
\end{align*}
To find the $(\xp, \yp) \in \sFb$ that maximizes the loss $\ell(\theta; \xp, \yp)$,
we therefore enumerate over the possible choices of $\yp$, solving the above convex problem for each $\yp$,
and pick the one that gives the smallest (most negative) margin.

\citet{steinhardt2017certified} show that if \eqref{eq:mm} is minimized incrementally
via $\epsilon n$ iterations of gradient descent, then we can form a strong attack $\sDp$ out of the corresponding set of $\epsilon n$
maximizers $\{(\xp,\yp)\}$.
Pseudocode is given
in Algorithm~\ref{alg:mm}.

\begin{algorithm}
\caption{\mmbasic{} attack.}
\label{alg:mm}
\begin{algorithmic}
\STATE Input: clean data $\sDc$, poisoned fraction $\epsilon$, burn-in $\nb$, feasible set $\sF_\beta$.
\STATE Initialize: $\theta \in \bR^d$, $\sDp \gets \emptyset$.
\FOR{$t = 1, \ldots, \nb + \epsilon n$}
  \STATE $Pick (\xp,\yp) \in \argmax_{(x,y) \in \sF_\beta} \ell(\theta; x,y)$.
  \hspace{5mm} (Find highest-loss point in $\sF_\beta$)
  \STATE $\theta \gets \theta - \eta (\lambda \theta + \nabla L(\theta) + \epsilon \nabla \ell(\theta; \xp, \yp))$
  \hspace{6.3mm} (Gradient update)
  \IF{$t > \nb$}
    \STATE $\sDp \gets \sDp \cup \{(\xp,\yp)\}$
    \hspace{26.4mm} (Add point to attack set)
  \ENDIF
\ENDFOR
\STATE Output $\sDp$.
\end{algorithmic}
\end{algorithm}

This algorithm automatically handles class balance,
since at each iteration it chooses to add either a positive or negative point;
it can thus handle multi-class attacks without additional difficulty,
unlike the \kkt{} or \inf{} attacks. Moreover, unlike the \inf{} attack,
it avoids solving the expensive bilevel optimization problem.

\subsubsection{Improvements to the basic algorithm}\label{sec:mm-improvements}
We improve the \mmbasic{} attack by incorporating the decoy parameters
introduced in \refsec{kkt} (\refalg{decoy}).
The problem with the \mmbasic{} attack, which repeatedly adds the highest-loss point that lies in the feasible set $\sF_\beta$, is that at low $\epsilon$,
the attack might end up picking poisoned points $\sDp$ that are not fit well by the model, i.e., with high $L(\hat\theta; \sDp)$.
These points could still lead to a high combined loss $L(\hat\theta; \sDcp)$,
but such an attack would be ineffective for two reasons:
\begin{enumerate}
\item If the poisoned points have high loss compared to the clean points,
  they are likely to be filtered by the \Loss{} defense.
\item Even if the poisoned points are not filtered out, the loss on the clean data $L(\hat\theta; \sDc)$ might still be low,
implying that the test loss would also be low.
Such a scenario could happen if there is no model that fits both the poisoned points $\sDp$ and the clean points $\sDc$ well;
since $\epsilon$ is small, overall training loss could then be minimized by fitting $\sDc$ well at the expense of $\sDp$.
\end{enumerate}

We therefore want to keep the loss on the poisoned points, $L(\hat\theta; \sDp)$, small.
To do so, we use the \emph{decoy parameters} $\thetadecoy$ from \refsec{kkt} (\refalg{decoy}),
augmenting the feasible set $\sF_\beta$ with the constraint
\begin{align}
\label{eqn:mm-loss-constraint}
\ell(\thetadecoy; x, y) \leq \tau,
\end{align}
for some fixed threshold $\tau$.
At each iteration, the attacker thus searches for poisoned points $(\xp, \yp)$ that maximize loss under the current parameters $\theta$
while having low loss under the decoy parameters $\thetadecoy$.
This procedure addresses the two issues above:
\begin{enumerate}
  \item If the learned parameters are driven towards $\thetadecoy$,
        the poisoned points in $\sDp$ will have low loss due to the constraint \refeqn{mm-loss-constraint},
        and hence will not get filtered by the \Loss{} defense.
\item Adding poisoned points with high loss under the current parameters but
      low loss under the decoy parameters $\thetadecoy$ is likely to drive the learned
      parameters towards $\thetadecoy$. In turn, this will increase the test loss,
      since $\thetadecoy$ is chosen to have high test loss \refsec{kkt-choosing-decoys}.
\end{enumerate}

We find that empirically, the \mmbasic{} attack naturally yields attacks that are quite concentrated.
For datasets with integer input constraints, we additionally use the linear programming relaxation and repeated points heuristic
(\refsec{improved-constraints}). Altogether, these changes to the \mmbasic{} attack yield what we call the \mm{} attack.

%% file: experiments.tex
\section{Experiments: Attackers with complete information}
\label{sec:experiments}

\subsection{Datasets}\label{sec:datasets}
Our experiments focus on two binary classification datasets. Summaries of each dataset are given in \reftab{data-characteristics},
including the number of training points $n$,
the dimension of each point $d$,
and the base accuracy of an SVM trained only on the clean data.
\begin{enumerate}
\item The Enron spam classification text dataset \citep{metsis2006spam}, which requires input to be non-negative and integer valued, as each feature represents word frequency.
The Enron dataset has $n \approx d$ and a relatively low base error of 3.0\%.
\item The IMDB sentiment classification text dataset \citep{maas2011imdb}, which similarly requires input to be non-negative and integer valued. Compared to the other datasets, the IMDB dataset has significantly larger $n$ and $d$, presenting computational challenges. It also has $n \ll d$ and is not as linearly separable, with a high base error of 11.9\%.
\end{enumerate}
In addition, we use the standard 10-class MNIST dataset \citep{lecun1998gradient} as an illustration of a multi-class setting.

These datasets are considered in \citet{steinhardt2017certified}, which also studied data poisoning. In \refapp{experiments-mnist-dogfish}, we also consider experiments on the other two datasets studied in that work: \mnist{}, a binary version of MNIST \citep{lecun1998gradient}, and Dogfish \citep{koh2017understanding}.
These datasets were shown by \citet{steinhardt2017certified} to have some certificates of defensibility using the \Ltwo{} and \slab{} defenses, and indeed, our attacks were not as effective on them as they were for the other datasets above.

\begin{table*}[h]
\centering
\begin{tabular}{l|r|r|r|r|r|l}
Dataset & Classes & $n$ & $\ntest$ & $d$ & Base Error & Input Constraints \\
\hline
Enron    & $2$  & $4137$  & $1035$  & $5116$  & $2.9\% \ (\lambda = 0.09)$ & $\bZ_{\geq 0}$ \\
IMDB     & $2$  & $25000$ & $25000$ & $89527$ & $11.9\% \ (\lambda = 0.01)$& $\bZ_{\geq 0}$ \\
MNIST    & $10$ & $60000$ & $10000$ & $784$   & $7.5\%$ & $[0,1]$\\
\mnist{} & $2$  & $13007$ & $2163$  & $784$   & $0.7\% \ (\lambda = 0.01)$ & $[0,1]$ \\
Dogfish  & $2$  & $1800$  & $600$   & $2048$  & $1.3\% \ (\lambda = 1.10)$ & $\bR$ \\
\end{tabular}
\caption{Characteristics of the datasets we consider, together with the base test errors that an SVM achieves on them (with regularization parameters $\lambda$ selected by validation). For multi-class MNIST, we used a multi-class SVM formulation \citep{crammer2002learnability}
with no explicit regularization and optimized it with AdaGrad \citep{duchi10adagrad}, which provides implicit regularization.
The input covariates for Enron and IMDB must be non-negative integers.}
\label{tab:data-characteristics}
\end{table*}

\subsection{Setup}

We used the non-iterative version of \refalg{constrained-opt} to carry out the data poisoning attacks.
We assumed that the attacker has limited control over the training data:
in our experiments, we allowed the attacker to only add up to $\epsilon=3\%$ poisoned data, and we set the data sanitization threshold $\tau$ such that the defender removes $p=5\%$ of the training data from each class after training its anomaly detector on the combined clean and poisoned dataset $\sDcp$.

As the attacker's goal is to increase test error regardless of which defense is deployed against it, we evaluated an attack $\sDp$ by running each of the 5 defenses in \refsec{defenses} separately against it,
and measuring the minimum increase in test error it achieves over all of the defenses.

We optimized each attack against all of the defenses.
Specifically, for the \inf{} attack, we took the feasible set $\sF_\beta$ to be the intersection of the feasible sets under the \Ltwo{} and \slab{} defenses,
plus any additional input constraints that each dataset imposed.
For the \kkt{} and \mm{} attacks, we used the decoy parameters to expand the feasible set $\sF_\beta$ to incorporate the \Loss{} defense as well.
We relied on concentrated attacks to evade the remaining defenses.

For the binary classification tasks, we used support vector machines (SVMs) with the hinge loss $\ell(\theta; x, y) = \max(0, 1 - y \theta^{\top}x)$
and with $L_2$ regularization, fixing the regularization parameter via
cross-validation on the clean data.\footnote{%
In practice, the defender does not have access to the clean data,
so its regularization parameter will be chosen based on the full dataset $\sDcp$.
However, our framework assumes that the attacker knows the regularization parameter in advance.
This is a potential disadvantage for the defender. In \refsec{transfer-reg}, we study what happens if the attacker does not know exactly how much regularization the defender will use.
}
For the multi-class task, we used the multi-class SVM formulation in \citet{crammer2002learnability}.
Further implementation details of our attacks are in \refapp{app_attack_details}.

\subsection{Comparing attacks on Enron and IMDB}

\begin{figure}[t]
\begin{center}
\includegraphics[width=1.0\columnwidth]{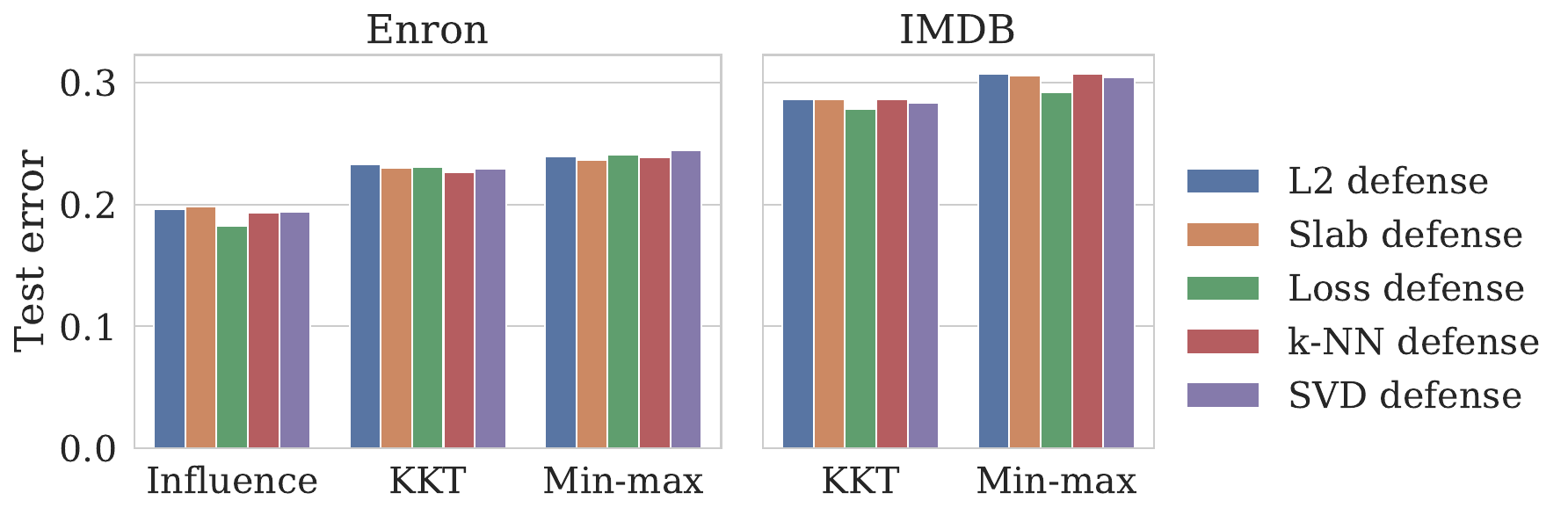}
\end{center}
\vspace{-3mm}
\caption{The \kkt{} and \mm{} attacks give slightly higher test error than the \inf{} attack on the Enron dataset. Moreover, they are more computationally efficient, and can be run on the larger and higher-dimensional IMDB dataset.}
\label{fig:all-test-error}
\end{figure}

We tested all three attacks on the Enron spam classification dataset, and the \kkt{} and \mm{} attack on the IMDB sentiment classification dataset (which was too large to run the \inf{} attack on).
All of the attacks were successful, with the \kkt{} and \mm{} attack achieving slightly higher test error than the \inf{} attack on the Enron dataset (\reffig{all-test-error}-Left).
As each defense is evaluated separately against the attack, we plot a bar for each defense.
However, our attacks are constructed to avoid all of the defenses;
this simulates the fact that the attacker might not know which defense will be deployed ahead of time, and therefore strives to evade all of them.

For the \kkt{} and \mm{} attacks, successful attacks did not need to exactly reach the decoy parameters;
in fact, trying to get to ambitious (i.e., high test error but unattainable) decoy parameters
could sometimes outperform exactly reaching unambitious decoy parameters.
(Of course, the ideal choice of decoy parameters would have high test error but also be attainable.)

\paragraph{Timing.}

\begin{figure}[t]
\begin{center}
\includegraphics[width=0.7\columnwidth]{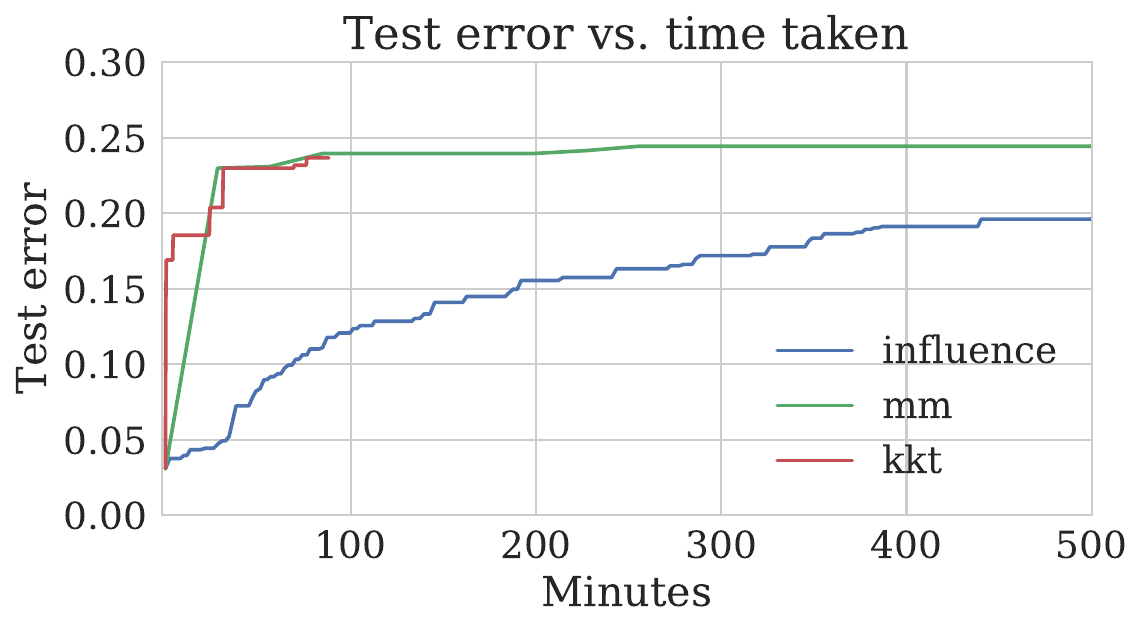}
\end{center}
\vspace{-5mm}
\caption{The test error achieved by the different attacks (against the \Ltwo{} defense), vs. the number of minutes taken to generate the attacks.
Each step increase in test error represents the processing of one choice of decoy parameters (for the \kkt{} and \mm{} attacks) or 10 gradient steps (for the \inf{} attack).}
\label{fig:timing}
\end{figure}

We measured the speed of each attack against the Enron dataset by running them each on 2 cores of an Intel Xeon E5 processor.
Additionally, the \inf{} attack used a GeForce GTX Titan X for GPU-based calculations.
Despite not using a GPU, the \kkt{} and \mm{} attacks were significantly faster:
while the \inf{} attack took 286 minutes to reach 17\% error, the \kkt{} attack only took 27 seconds (\reffig{timing}).
The \mm{} attack took 28 minutes to process its first decoy parameter, which gave 23.0\% error.
The main computational bottleneck for the \inf{} attack is the inverse Hessian-vector product calculation in Equation \refeqn{influence-params}.
In contrast, the other two attacks solve convex subproblems (as opposed to the non-convex problem that the \inf{} attack is doing gradient descent on) and admit more efficient general-purpose convex optimization solvers.

\subsection{Iterative optimization}\label{sec:expt_iterative}

\begin{figure}[t]
\begin{center}
\includegraphics[width=0.8\columnwidth]{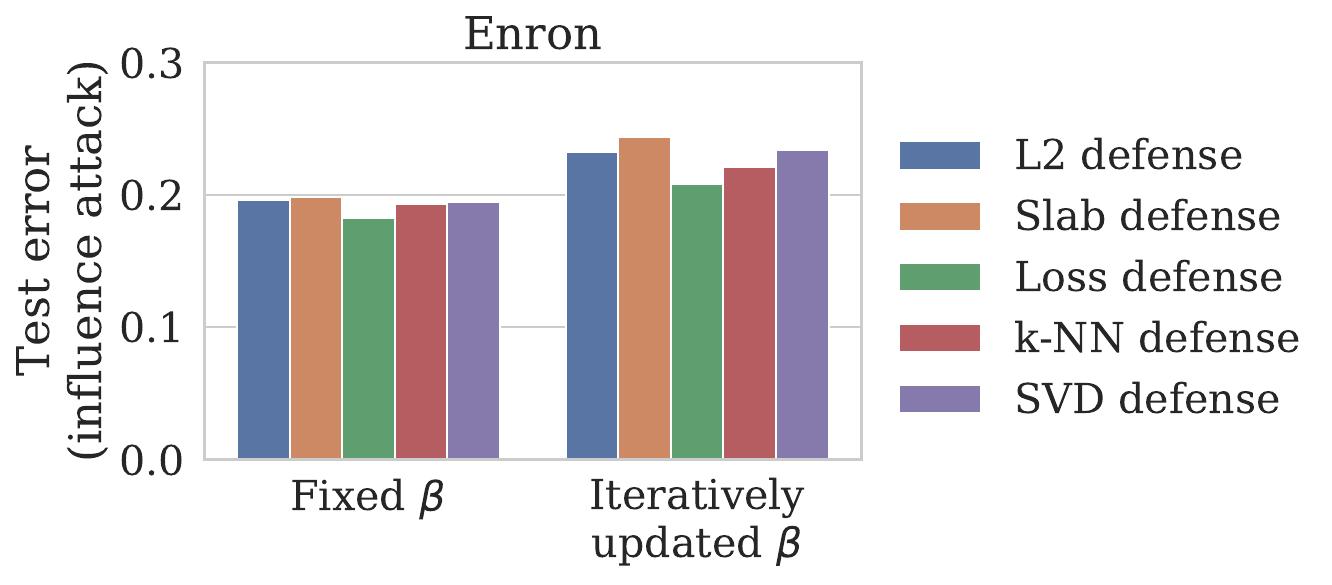}
\end{center}
\vspace{-5mm}
\caption{Iteratively updating the feasible set $\sF_\beta$ increases test error by a few percentage points on the Enron dataset (with $\epsilon = 3\%$ poisoned data), compared to fixing the feasible set
based on just the clean data $\sDc$.}
\label{fig:influence-em}
\end{figure}

The above experiments simply fix the feasible set $\sF_\beta$ based on the clean training data, as described in \refsec{optimization}.
To study the effect of refining the attacks by iteratively $\sF_\beta$ (\refalg{constrained-opt}),
we ran an iterative version of the \inf{} attack against the Enron dataset.
\reffig{influence-em} shows that iterative optimization does only slightly better:
at the low levels of $\epsilon$ that we chose, the attack does not shift the centroids of the data that much,
and therefore the feasible set $\sF_\beta$ for both the \Ltwo{} and \slab{} defenses stay somewhat constant.

One perspective on iterative optimization in our setting is that it is targeting the \slab{} defense by trying to rotate the vector between the two class centroids;
as \citet{steinhardt2017certified} show, at large $\epsilon$ (e.g., $\epsilon = 0.3$, which is 10 times larger than what we consider), this vector can be significantly changed by the poisoned points,
whereas the $\Ltwo{}$ feasible set is harder to perturb.\footnote{
Note that our \inf{} attacks on \mnist{} at high $\epsilon$ are considerably weaker than the attacks in \citet{steinhardt2017certified},
which uses a specialized semi-definite program that relies on poisoning the anomaly detector (i.e., placing poisoned points to move the class centroids in a way that renders the \slab{} defense ineffective).
They achieve an increase in test error of 39\% for $\epsilon = 0.3$.
The high-$\epsilon$ setting is not our focus in this paper, since it is less realistic;
this performance gap could be a result of poor step size tuning or initialization on our part,
or it could signify a weakness in the applicability of iterative optimization and/or gradient descent to the high-$\epsilon$ setting.
}
On the Enron dataset and at the low $\epsilon$ settings that we consider, the slab defense only decreases test error by a few percentage points, so iterative optimization only increases test loss by a few percentage points.
To illustrate this point, we ran an attack with $\epsilon = 0.3$ on the \mnist{} dataset:
the \inf{} attack without iterative optimization achieved an increase in test error of only 1.1\%,
while the \inf{} attack with iterative optimization achieved a larger increase in test error of 7.5\%.

\subsection{Ablations for the \inf{} attack}
\begin{figure}[h]
\begin{center}
\includegraphics[width=\columnwidth]{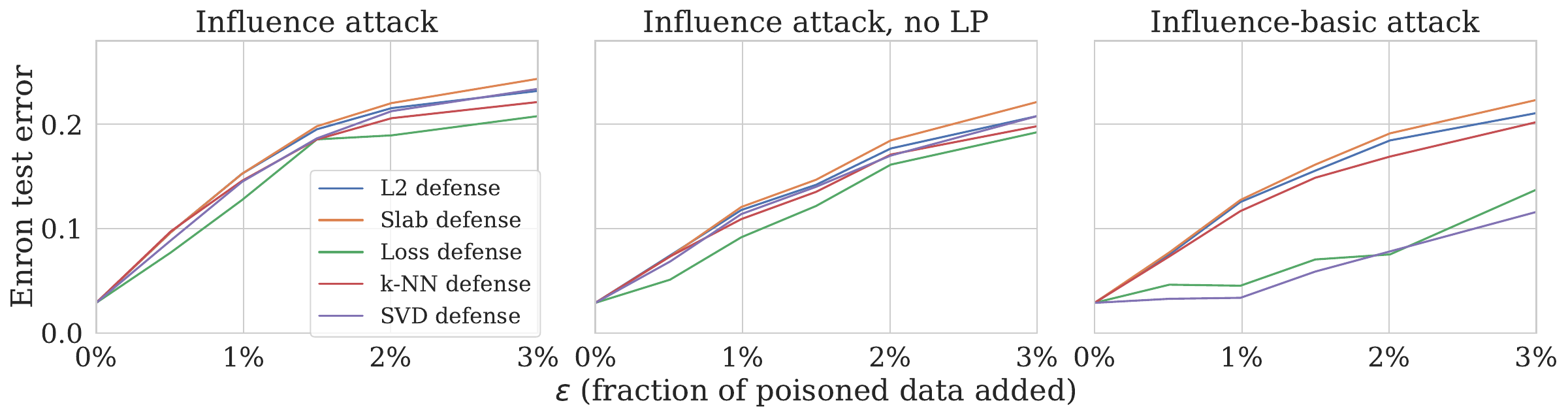}
\end{center}
\vspace{-5mm}
\caption{Ablative study of the changes we made to the \inf{} attack, evaluated on the Enron dataset.
Left: results of the \inf{} attack.
Middle: results of the \inf{} attack, without the linear programming (LP) relaxation described in \refsec{improved-constraints}.
Right: results of the \infbasic{} attack, which does not use the LP relaxation nor a concentrated attack.
}
\label{fig:influence-ablative}
\end{figure}

We studied the effect of the two improvements made to the \infbasic{} attack---the linear programming (LP) relaxation and the concentration of the attack---on the Enron dataset.
Removing the linear programming (LP) relaxation decreased the achieved test error by a few percentage points (\reffig{influence-ablative}-Left versus Mid).
Further removing the concentrated attack decreased test error substantially (\reffig{influence-ablative}-Right).
The \infbasic{} attack is still optimized to evade the \Ltwo{} and
\slab{} defenses, but because its poisoned points are not concentrated,
many of them get filtered out by the \Loss{} and \PCA{} defenses.
Consequently, it does not manage to increase test error beyond 11\% under those defenses.

\subsection{Ablations for the \mm{} attack}
\begin{figure}
\begin{center}
\includegraphics[width=0.8\columnwidth]{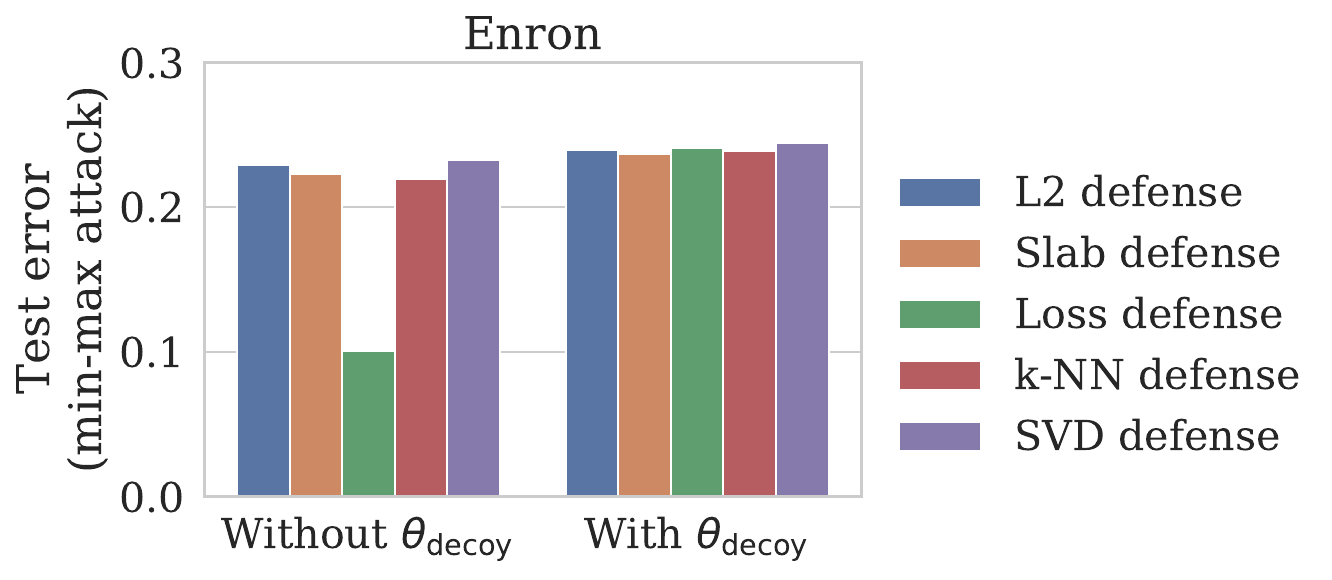}
\end{center}
\vspace{-5mm}
\caption{The use of decoy parameters allows the \mm{} attack to evade the \Loss{} defense.}
\label{fig:mm-ablative}
\end{figure}

To measure the effect of using decoy parameters in the \mm{} attack,
we ran the \mmbasic{} attack from \citet{steinhardt2017certified}, augmented with the linear programming relaxation and repeated points heuristic.
(The unaugmented version of the \mmbasic{} attack from \citet{steinhardt2017certified} performs worse.)
In contrast to the \mm{} attack, the \mmbasic{} attack gets defeated by the \Loss{} defense (test error 10.0\%, \reffig{mm-ablative}).
As discussed in \refsec{mm-improvements}, without the constraints imposed by the decoy parameters, the poisoned points found by the \mmbasic{} attack have high loss and consequently get filtered out by the \Loss{} defense.

\subsection{Attacks on multi-class tasks with the \mm{} attack on MNIST}

Finally, one advantage of the \mm{} attack is that it works on a variety of input domains $\sX \times \sY$ and non-convex feasible sets $\sF_\beta \subset \sX \times \sY$, so long as we can still efficiently solve $\max_{(x, y) \in \sF_\beta} \ell(\theta; x, y)$.
In particular, for multi-class problems, the \mm{} attack can search over different choices of $\yp$ for each poisoned point $(\xp, \yp)$, whereas the \inf{} and \kkt{} attacks require us to grid search over the relative proportion of the different classes.
As we need $k(k-1)$ distinct points to carry out any data poisoning attack against a $k$-class SVM (\refprop{kpoints} in Appendix A), this grid search would take time exponential in $k^2$.

We illustrate a multi-class attack by running the \mm{} attack on the 10-class MNIST dataset \citep{lecun1998gradient}.
Using $\epsilon = 3\%$ poisoned data, the \mm{} attack obtains 15.2\% test error against the \Ltwo{} defense and 13.7\% test error against the \Loss{} defense, demonstrating a high-leverage attack in a multi-class setting.

\begin{figure}[h]
\begin{center}
\includegraphics[width=0.6\columnwidth]{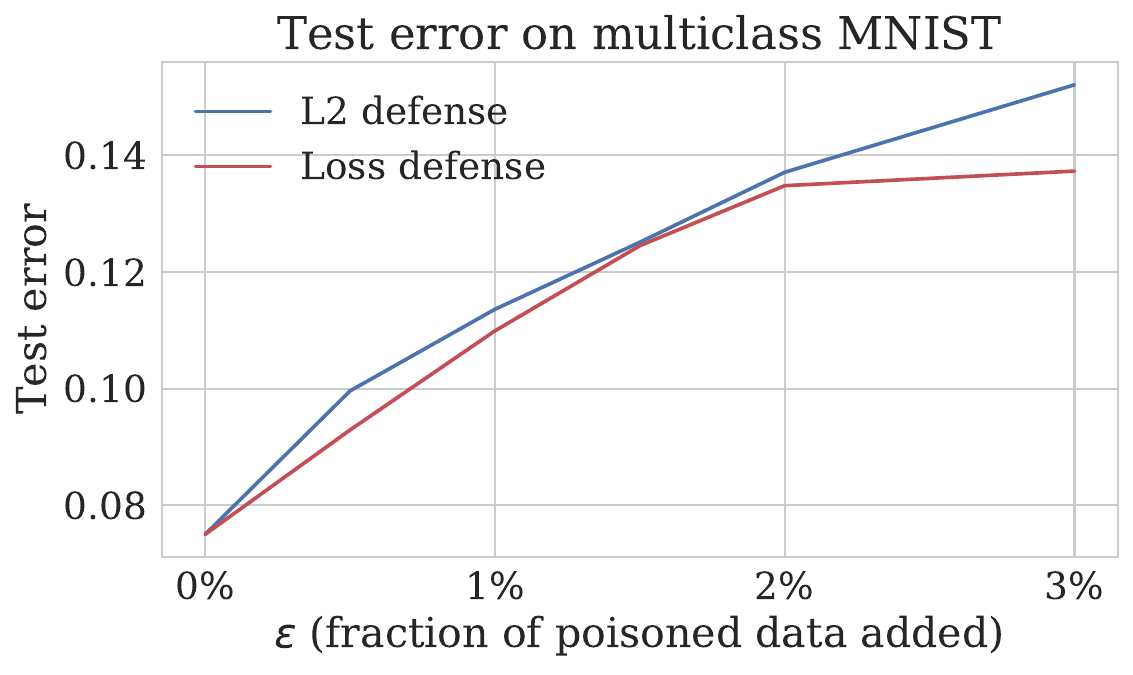}
\end{center}
\vspace{-5mm}
\caption{The \mm{} attack scales to handle multi-class attacks, as it automatically chooses
the class balance / relative proportion of poisoned points. Here, we show that
the \mm{} attack can increase test error on the MNIST dataset to 13.7\% with $\epsilon=3\%$ poisoned data.}
\label{fig:mm-multiclass}
\end{figure}

%% file: transfer.tex
\section{Experiments: Attackers with incomplete information}
\label{sec:transfer}

In \refsec{experiments}, we saw that the \inf{}, \kkt{}, and \mm{} attacks are effective if the attacker has complete information about the model and defenses.
In this section, we study transferability---are these attacks still effective when the attacker does not have complete information about the defender?
Specifically, we use the Enron dataset to explore what happens when the attacker does not have knowledge of 1) the test set (i.e., they only see the train set); 2) the amount of regularization that the defender uses; 3) their optimization algorithm; and 4) their loss function.

In general, our attacks are still effective under these changes, with the \mm{} attack generally being the most robust and the \inf{} attack being the least.
However, the \Loss{} defense poses problems for all three attacks when the optimization algorithm or the loss function are changed, suggesting that attackers should set conservative thresholds against the \Loss{} defense.

\subsection{Unobserved test data}\label{sec:transfer-train}
An attacker might not have specific test examples in mind; for example, they might aim to make the defender incur high expected loss on test points from the same distribution as the training set. For such an attacker, optimizing for some test data $\sDtest$ might not translate into an effective attack on a different sample of test data.
We tested if our attacks would still be effective if the attacker only knew the clean training data $\sDc$ but not the test data $\sDtest$.\footnote{
A different setting is if the attacker knows the test data $\sDtest$ but not the clean training data $\sDc$. For example, the attacker might only have access to a similarly distributed but distinct dataset $\sDc'$. As our attacks depend on the clean training data primarily through the average gradient of the loss computed over $\sDc$, we expect that swapping $\sDc$ with $\sDc'$ should not matter for sufficiently large training sets.}
Specifically, we generated attacks by simply using $\sDc$ in place of $\sDtest$ (i.e., we optimized for high error on the training data $\sDc$).

\begin{figure}[h]
\begin{center}
\includegraphics[width=\columnwidth]{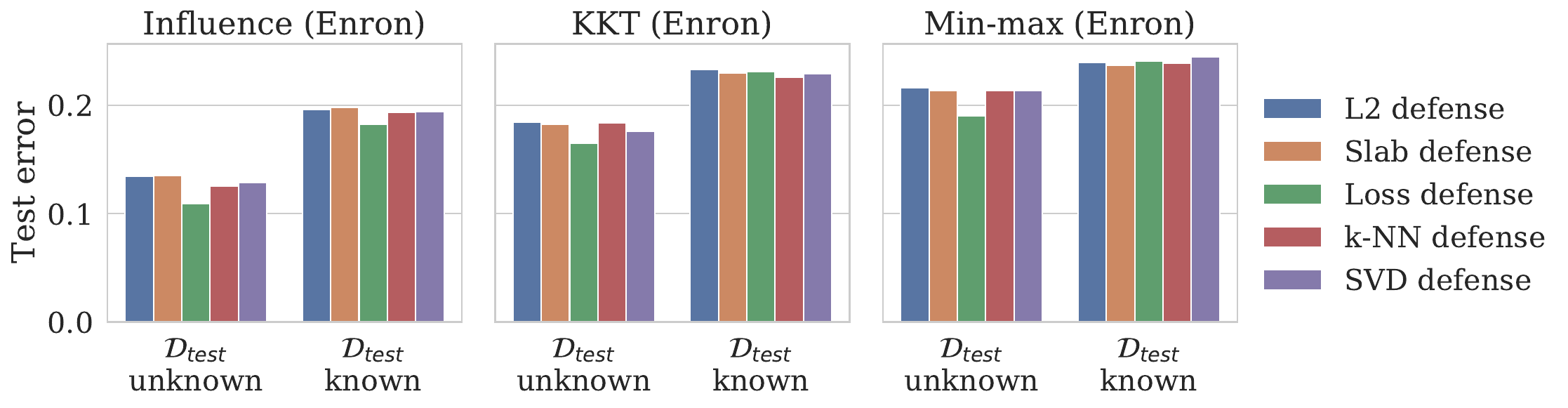}
\end{center}
\vspace{-5mm}
\caption{The performance of each attack when the test data $\sDtest$ is unknown and the attack is optimized against the training data (left columns) vs.~in the setting where $\sDtest$ is known (right columns).}
\label{fig:transfer-train}
\end{figure}

\reffig{transfer-train} shows the test error that the resulting attacks incurred on $\sDtest$, compared with attacks that assume knowledge of $\sDtest$, as used in the rest of the paper.
Overall, the attacks still significantly increase test error even without knowing $\sDtest$. However, the \inf{} attack is comparatively less effective when the test dataset is not known (reaching 10.9\% test error instead of 18.3\% test error). In contrast, the \kkt{} (16.5\% vs. 22.6\% test error) and \mm{} attacks (19.0\% vs 23.7\% test error) are more robust to using the training data in place of the test data. These results suggest that the \inf{} attack, which explicitly optimizes for loss on the test set, overfits more strongly to the test set compared to the \kkt{} and \mm{} attacks, which rely on the test set mainly through the construction of decoy parameters.

A variant of the above setting is when the attacker does not know the exact test data $\sDtest$ but nonetheless has access to some validation data $\sD_\text{val}$ from the same distribution, as well as the training data $\sDc$. In this setting, the attacker could optimize an attack against $\sD_\text{val}$; we expect that doing so will result in an attack that is more effective than optimizing against the training data $\sDc$, as we do above, though still slightly less effective than optimizing against $\sDtest$ itself.

\subsection{Regularization}\label{sec:transfer-reg}

\begin{figure}[t]
\begin{center}
\includegraphics[width=0.8\columnwidth]{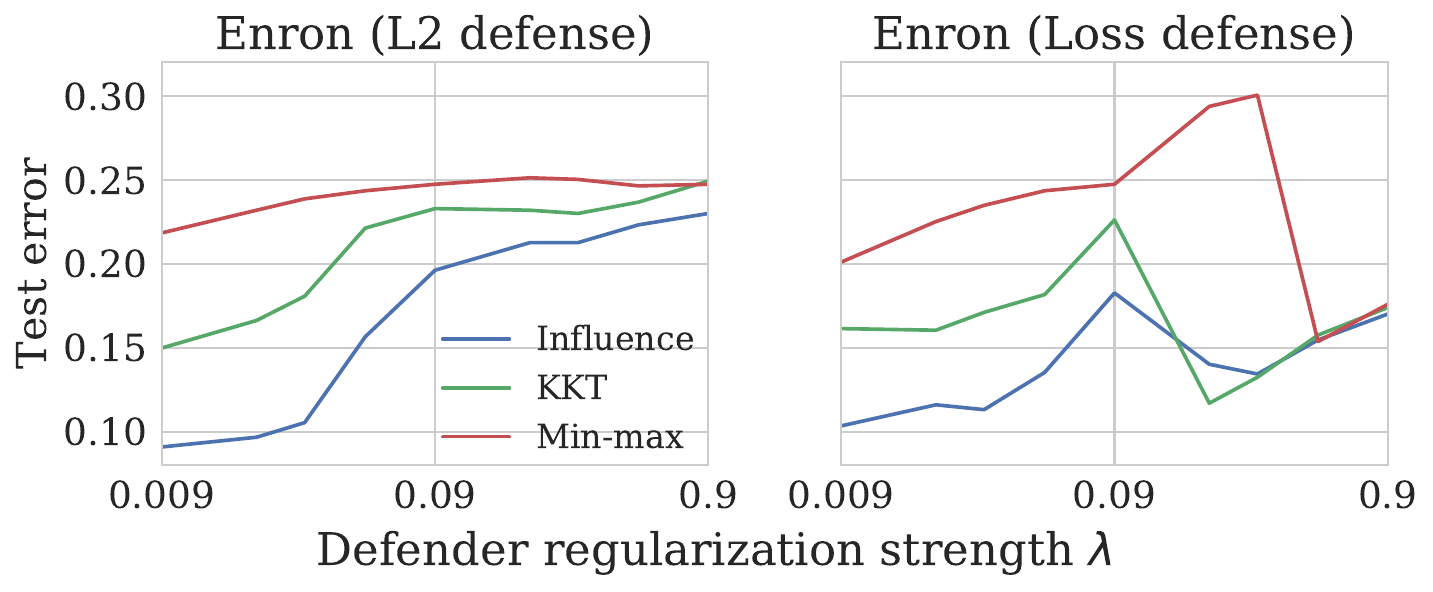}
\end{center}
\vspace{-5mm}
\caption{The effect of changing the defender's regularization strength on
the test error achieved by each attack. The attacker's regularization
is fixed at $\lambda = 0.09$.}
\label{fig:transfer-wd}
\end{figure}

Do attacks that are optimized for one level of regularization still work well at other levels of regularization?
Recall that the defender uses $L_2$ regularization with the hyperparameter $\lambda$ controlling the amount of regularization (Equation \refeqn{hattheta}); in particular, we use $\lambda=0.09$ for the Enron dataset (\reftab{data-characteristics}).
To test the effect of the defender's choice of $\lambda$, we varied it from $0.009$ to $0.9$ while keeping the attacker's $\lambda$ constant at $0.09$.
\reffig{transfer-wd} shows the results:
\begin{itemize}
\item Against the \Ltwo{} defense, test error generally increased with $L_2$ regularization strength (\reffig{transfer-wd}-Left).
The \Ltwo{} feasible set depends only on the location of the poisoned points and not on the model parameters that the defender learns.
Thus, the test error changes because the defender learns a different set of parameters given exactly the same set of poisoned points, and not because a different set of poisoned points get filtered out by the defenses.

\item The amount of regularization had different effects on each attack's effectiveness against the \Ltwo{} defense. The \inf{} attack became less effective as we reduced defender regularization ($<10\%$ test error against the \Ltwo{} defense at $\lambda=0.009$), while the \mm{} attack was robust to changes in defender regularization ($22\%$ test error at $\lambda=0.009$).

\item Increasing regularization can make the \Loss{} defense more effective (\reffig{transfer-wd}-Right).
  The \Loss{} feasible set is sensitive to changes in the model parameters that the defender learns,
 so some poisoned points that evaded this defense under the original model (with $\lambda=0.09$) are now detected under the changed model.

 \item Unlike the other attacks, the \mm{} attack initially gets more effective against the \Loss{} defense as defender regularization
 is increased from $\lambda=0.09$. We suspect that this is due to the \mm{} attack using a fixed loss threshold $\tau$ (see equation \refeqn{mm-loss-constraint} in \refsec{mm-improvements})
 that is more conservative than the \kkt{} attack (which uses an adaptive threshold based on the quantiles of the loss under the decoy parameters) and the \inf{} attack (which solely relies on concentrated attacks to overcome the \Loss{} defense).

\end{itemize}

These results imply that attackers should optimize for lower levels of regularization, in case the defender uses a lower level (and conversely, it suggests that defenders might want to use lower levels of regularization than they might otherwise).
Attackers using decoy parameters might also decide to set their loss thresholds more conservatively.

These results also suggest that the data poisoning attacks are not exploiting overfitting. If that were the case, we would expect increasing regularization to decrease overfitting and thus reduce attack effectiveness.
Instead, we observe the opposite: the defender generally suffers when increasing regularizatio, as it is harder for the defender's model to fit both the poisoned training points and clean training points well, and if the clean training points are not fit well, the test error will consequently increase.

We note that \citet{demontis2019adversarial} studied the transferability of gradient-based data poisoning attacks for image recognition and found that attacks were more effective when the defender used less regularization. The differences from our work is that they assumed that there are no defenses (i.e., the only constraint on the attacker is to generate a valid image) and that there are a small number of training points relative to dimension (e.g., $n=500$ for a binary MNIST classification problem).

\subsection{Optimization and loss function}
In our previous experiments, we assumed that the defender would learn the model parameters $\hat\theta$ that globally minimized training loss. In practice, defenders might use stochastic optimization and/or early stopping, leading to parameters $\hat\theta$ that are close to but not exactly at the optimum.

\begin{figure}[h]
\begin{center}
\includegraphics[width=1.0\columnwidth]{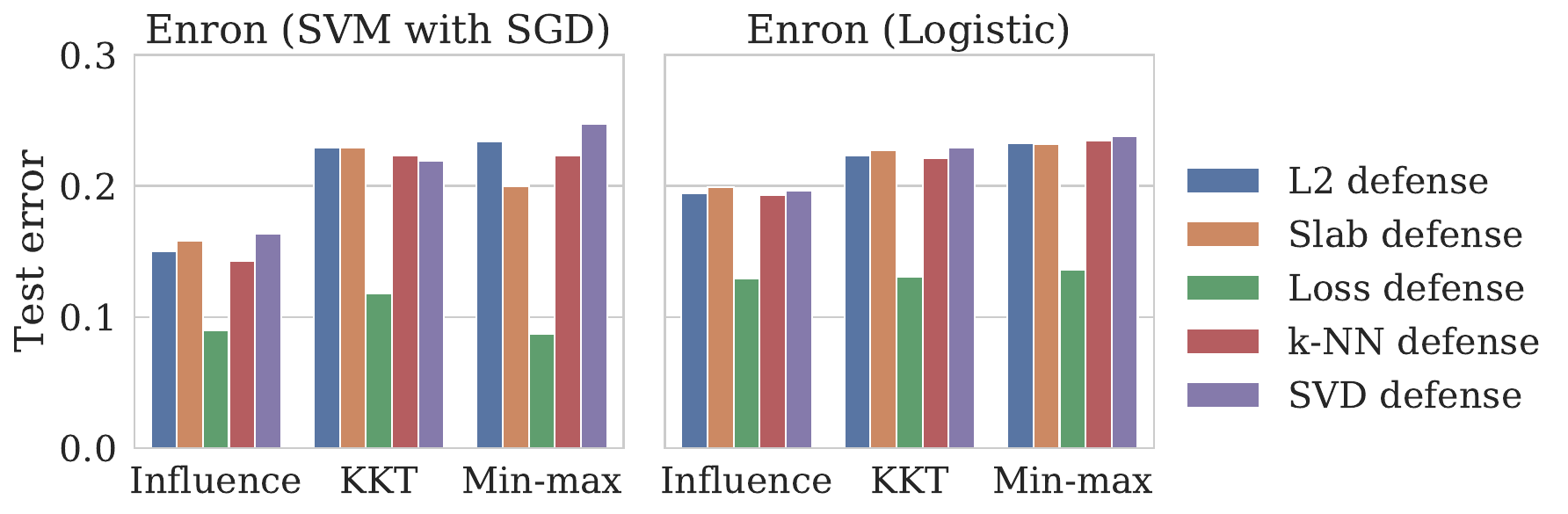}
\end{center}
\vspace{-5mm}
\caption{With the significant exception of the \Loss{} defense, the attack results are robust to
shifts in the optimization algorithm (from an exact solution to a single pass of stochastic gradient descent) and the loss function (from the hinge loss to the logistic loss).}
\label{fig:transfer-sgdlog}
\end{figure}

\reffig{transfer-sgdlog}-Left shows the results of our attacks on a defender that learned a model by doing a single pass of stochastic gradient descent over the dataset (i.e., each sample is looked at exactly once).
We also tested how an attacker using the hinge loss would fare against a defender who uses the logistic loss (\reffig{transfer-sgdlog}-Right).
All three attacks stayed effective against all of the defenses except the \Loss{} defense, which managed to significantly reduce the damage inflicted by the attacker.

As in \refsec{transfer-reg}, these results suggest that attackers should use a conservative loss threshold to harden their attacks against loss-based defenses.
It also suggests that the attacker's ability to evade loss-based defenses is especially sensitive (relative to other defenses) to getting the defender's model correct.

%% file: related.tex
\section{Related Work}\label{sec:related}

In this section, we discuss other attack settings and defense strategies that have been studied in the literature.
For broad surveys on this topic, see e.g., \citet{barreno2010security}, \citet{biggio2014security},  \citet{gardiner2016security},  \citet{papernot2016towards}, and \citet{vorobeychik2018adversarial}.

\subsection{Attacks}

\paragraph{Label-flip attacks.} The attacks presented in this work control both the label $\yp$ and input features $\xp$ of the poisoned points.
Attacks that only control the label $\yp$ are known as \emph{label-flip attacks}.
In a typical label-flip attack, the attacker gets to change some $\epsilon$ fraction
of the labels $y$ of the training set but is unable to change the features $x$ \citep{biggio2011label, xiao2012adversarial, xiao2015contamination}.
We experimented with a variant of the label flip attack described in \citet{xiao2012adversarial},
where we allowed the attacker to pick examples from the test set, flip their labels, and add them to the training set (\refapp{alfa}).
We found that this attack, though fast to run, was significantly less effective than our proposed attacks; control of $\xp$ seems to be necessary to carry out high-leverage attacks in the presence of data sanitization defenses.

\paragraph{Targeted vs. indiscriminate attacks.}
In our experiments, the attacker sought to increase error on a test set $\sDtest$ that was drawn from the same distribution as the clean training data $\sDc$.
This type of attack is known as an \emph{indiscriminate} attack \citep{barreno2010security} and is akin to a denial-of-service attack.

Indiscriminate attacks seek to change the predictions of the learned model on a good fraction of the entire data distribution
and therefore require substantial changes to the model.
This makes indiscriminate attacks statistically interesting,
as they get at fundamental properties of the model:
how might an attacker that only controls 1\% of the training data
bring about a 10\% increase in test error?

A different type of attack is a \emph{targeted} attack, in which the attacker seeks to cause errors on specific test examples or small sub-populations of test examples
\citep{gu2017badnets, chen2017targeted, burkard2017analysis, koh2017understanding, shafahi2018poison, suciu2018does}.
For example, an attacker might seek to have all of the emails that they send marked as non-spam while leaving other emails unaffected;
or an attacker might seek to cause a face recognition system to recognize their face as that of a particular victim's (as in \citet{biggio2012biometric} and \citet{biggio2013biometric}, which build off the attacks in \citet{kloft2012security}).
Targeted attackers only seek to change the predictions of the model on a small number of instances,
and therefore might be able to add in 50 poisoned training points to cause an error on a single test point \citep{shafahi2018poison}.
Targeted attacks are well-motivated from a security perspective:
attackers might only care about a subset of the model's prediction,
and targeted attacks require less control over the training set and are therefore easier to carry out.

The \inf{} and \kkt{} attacks in \refsecs{improved-inf}{kkt} do not make any assumptions on the nature of the test set $\sDtest$,
and can therefore handle the targeted attack setting without modification.
In contrast, the \mm{} attack in \refsec{improved-mm} assumes that the training error is a good approximation to the test error,
and is thus only appropriate in the indiscriminate attack setting.

\paragraph{Backdoor attacks.}
A backdoor attack is a targeted attack that seeks to cause examples that contain a specific backdoor pattern, e.g., a bright sticker \citep{gu2017badnets} or a particular type of sunglasses \citep{chen2017targeted}, to be misclassified.
Backdoor attacks work by superimposing the chosen backdoor pattern onto particular training examples from a given class, which causes the model to associate the backdoor pattern with that class (and therefore misclassify, at test time, examples of a different class that also contain the backdoor pattern).
The attackers in \citet{gu2017badnets} and \citet{chen2017targeted} do not need to know the model that the defender is using; in fact, the attacker in \citet{chen2017targeted} does not even need any knowledge of the training set, instead adding examples from a external dataset.
These weaker assumptions on attacker capabilities are common in targeted attacks.
In contrast, the indiscriminate attacks that we develop in this paper make use of knowledge of the model and the training set in order to have high leverage.

\paragraph{Clean-label attacks.}
Clean-label attacks are attacks that ``do not require control over the labeling function; the poisoned training data appear to be labeled correctly according to an expert observer'' \citep{shafahi2018poison}.
Not requiring control over labeling makes it easier for the attacker to practically conduct such an attack,
as the attacker only needs to introduce the unlabeled poisoned data into the general pool of data
(e.g., uploading poisoned images or sending poisoned emails, as \citet{shafahi2018poison} discusses)
and wait for the defender to label and ingest the poisoned data.

The backdoor attacks discussed above are examples of clean-label attacks, provided the backdoor pattern is chosen to be innocuous enough to avoid human suspicion.
Another way of constructing clean-label attacks is by constraining the poisoned points to be close, in some metric, to a clean training point of the same class;
\citet{shafahi2018poison} does this with the $\ell_2$ norm, while \citet{suciu2018does} and \citet{koh2017understanding} use the $\ell_\infty$ norm.

Our attacks are not clean-label attacks, in that the poisoned points will not necessarily be labeled as the correct class by a human expert.
On the other hand, our poisoned points are designed to evade detection by automatic outlier detectors;
note that ``clean-label'' points can fool human experts but still look like statistical outliers.

\paragraph{Adversarial examples and test-time attacks.}
The bulk of recent research in machine learning security has focused on test-time attacks,
where the attacker perturbs the test example to obtain a desired classification,
leaving the training data and the model unchanged.
This line of research was sparked by the striking discovery that
test images could be perturbed in a visually-imperceptible way and yet fool
state-of-the-art neural network image classifiers
\citep{szegedy2014intriguing,goodfellow2015explaining,carlini2016hidden,
kurakin2016adversarial,papernot2016transferability,papernot2017practical,moosavi2016deepfool}.
Designing models that are robust to such attacks, as well as coming up with more effective attacks, is an active area of research
\citep{papernot2016distillation, madry2017towards,
tramer2017ensemble, wong2018provable,
 raghunathan2018certified, sinha2018certifiable, athalye2018obfuscated,
 papernot2018deep}.

In contrast to these test-time attacks, data poisoning attacks are train-time attacks:
the attacker leaves the test example unchanged, and instead perturbs the training data so as to affect the learned model.
Data poisoning is less well-studied, and compared to test-time attacks,
it is harder to both attack and defend in the data poisoning setting:
data poisoning attacks have to depend on the entire training set,
whereas test-time attacks only depend on the learned parameters;
and common defense techniques against test-time attacks, such as 'adversarial training'
\citep{goodfellow2015explaining}, do not have analogues in the data poisoning setting.

\subsection{Defenses}
In the literature, effective defenses typically require additional information than the defenders we consider in this paper,
e.g., having a labeled set of outliers or having a trusted dataset.

\paragraph{Using labeled outlier data and other training metadata.}
If the defender has access to data that has been labeled as `normal' vs. `outlier', then outlier detection can be treated as a standard supervised classification problem \citep{hodge2004survey}.
For example, an online retailer might have a set of transactions that had been previously labeled as fraudulent,
and could train a separate outlier detection model to detect and throw out other fraudulent-looking transactions from the dataset.
One drawback is that in an adversarial setting there is no assumption that future poisoned points might look like previous poisoned points.
Such methods are therefore more suited for detecting outliers caused by natural noise processes rather than adversaries.

Instead of directly using `normal' vs. `outlier' labels, defenders can instead rely on other types of training metadata.
For example, \citet{cretu2008casting}---which introduced the term `data sanitization' in the context of machine learning---uses information on the time at which each training point was added to the training set.
The intuition is that ``in a training set spanning a sufficiently large time interval, an attack or an abnormality will appear only
in small and relatively confined time intervals'' \citep{cretu2008casting}.

\paragraph{Using trusted data.}
Other defense methods rely on having a trusted subset $\sT$ of the training data that only contains clean data (obtained for example by human curation).
One example is the Reject on Negative Impact (RONI) defense proposed by \citet{nelson2008exploiting},
which was one of the first papers studying data poisoning attacks and defenses.
The RONI defense iterates over training points $(x, y)$
and rejects points if the model learned on just the trusted data $\sT$ is significantly different
from the model learned on $\sT \cup \{(x, y)\}$.
Another example is the outlier detector introduced in \citet{paudice2018detection},
which operates similarly to our \knn{} defense except that it measures distances only to the points in the trusted subset.

Having a trusted dataset makes it easier for the defender, though such a dataset might be expensive or even impossible to collect;
if the defender has enough resources to collect a large amount of trusted data, then they can train a model on only the trusted data, solving the problem of data poisoning.
The question of whether a small amount of trusted data is sufficient for defeating attackers while maintaining high overall performance
(i.e., not rejecting clean training points that are not similar to the trusted data) is an open one.
Finally, defenses that rely on trusted data are particularly vulnerable to attackers that manage to compromise the trusted data
(e.g., through clean-label attacks that escape human notice).

\paragraph{High-dimensional robust estimation.}
The theoretical computer science community has studied robust estimators in high dimensions, which seek to work well even in the presence of outliers \citep{kearns1993learning}.
A key issue is that many traditional robust estimators incur a
$\sqrt{d}$ increase in error in $d$ dimensions.
This theoretical insight aligns with the
empirical results in this paper showing that it is often possible to attack classifiers
with only a small fraction of poisoned data.

Motivated by these issues, \citet{klivans2009learning} and later
\citet{awasthi2014power} and \citet{diakonikolas2017learning}
design robust classification algorithms that
avoid the poor dimension-dependence of traditional estimators, although only
under strong distributional assumptions such as log-concavity.
Separately, \citet{lai2016agnostic} and \citet{diakonikolas2016robust} designed
robust procedures for mean estimation, which again required strong distributional
assumptions. Later work \citep{charikar2017learning,diakonikolas2017practical,steinhardt2018resilience} showed how to perform mean estimation under much more mild assumptions,
and \citet{diakonikolas2017practical} showed that their procedure could yield robust estimates
in practice. In recent concurrent work, \citet{diakonikolas2018sever} showed that robust
estimation techniques can be adapted to classification and used this to design a practical
algorithm that appears more robust than many traditional alternatives. It would
be interesting future work to attack this latter algorithm in order to better vet its robustness.

\paragraph{Certified defenses.}
\citet{steinhardt2017certified} explore the task of provably certifying defenses,
i.e., computing a dataset-dependent upper bound on the maximum test loss that an attacker can cause the defender to incur.
Their method---from which we adopt the \mmbasic{} attack---shows that the
\Ltwo{} and \slab{} defenses are sufficient for defending models trained on the \mnist{} and Dogfish datasets but cannot certifiably protect models trained on the Enron and IMDB datasets, which matches with our experimental results.
Open questions are whether our improvements to the \mmbasic{} (e.g., decoy parameters) can be used in their framework to derive tighter upper bounds on attack effectiveness, and whether the other defenses (e.g., the \Loss{} defense) can be incorporated into their framework.

%% file: discussion.tex
\section{Discussion}\label{sec:discussion}

In this paper, we developed three distinct attacks that could evade data sanitization defenses
while significantly increasing test error on the Enron and IMDB datasets.
The \inf{} attack directly optimizes for increasing the test loss through gradient ascent on the poisoned points;
the \kkt{} attack chooses poisoned points to achieve pre-determined decoy parameters;
and the \mm{} attack efficiently solves for the poisoned points that maximize train loss, as a proxy for test loss.

We summarize the relative merits of these attacks in \reftab{attack-comparison}.
The \inf{} attack is direct but slow, less effective against model-based defenses (such as the \Loss{} defense), and less robust. The \kkt{} attack is much faster, at least in our setting where it can heavily exploit convexity; however, its reliance on  decoy parameters is also a limitation, as our heuristic for generating decoy parameters might fail under more sophisticated defenses. The \mm{} attack shares the same reliance on decoy parameters and is slower than the \kkt{} attack, and it only works in the indiscriminate setting, but it is more robust and can handle multi-class settings more efficiently.

We expect that more sophisticated data sanitization defenses could defeat the attacks developed in this paper, which did not account for them. However, these defenses might in turn be broken by attacks that are specifically geared for those defenses.
The results in this paper show that data poisoning defenses need to be tested against attackers that are designed to evade them.
We end by discussing some directions for future work.

\newcommand{\yes}{\ding{51}}
\newcommand{\no}{\ding{55}}

\begin{table*}[t]
\centering
\begin{tabular}{lrrl}
\toprule
\bf{Attack} & \bf{Enron test error}& \bf{Time taken} & \bf{Pros and cons}\\
\midrule
Influence & $18.3\%$ & 17,000s & \makecell[l]{
  \yes~No need to find $\thetadecoy$\\
  \no~Slow\\
  \no~Less transferable\\
  \no~Does not handle \Loss{} defense}\\
\hline
KKT & $22.5\%$ & 27s  & \yes $\thetadecoy$ handles \Loss{} defense\\
\hline
Min-max & $23.7\%$ & 1,700s & \makecell[l]{
  \yes~$\thetadecoy$ handles \Loss{} defense\\
  \yes~Handles multi-class setting\\
  \no~Assumes indiscriminate attack}\\
\bottomrule
\end{tabular}
\caption{Comparison of attacks. Enron test error is reported at $\epsilon=3\%$,
and the time taken is how long each attack took to reach $17\%$ Enron test error.}
\label{tab:attack-comparison}
\end{table*}

\paragraph{What makes datasets and models attackable?} The effectiveness of our attacks vary significantly from dataset to dataset
(e.g., linear models trained on the Enron and IMDB datasets are more vulnerable than linear models trained on the \mnist{} and Dogfish datasets).
What conditions make certain models on certain datasets attackable, but not others?
This is an open question; we speculate that linear models on the Enron and IMDB datasets are easier to attack because they have higher dimensionality and are less linearly separable,
but this question deserves more study.

\paragraph{Non-convex models.}
One limitation of our attacks is that they rely on the attacker and defender being able to find the model parameters $\theta$ that globally minimize the training loss.
This is a reasonable assumption if the loss is convex, but most models in practice today are non-convex.
Analysis of attack algorithms in the non-convex setting is substantially more difficult, e.g.,
a poisoning attack that works against a neural net trained with a given random seed might fail when the random seed is changed,
or a single poisoned point might cause the defender to get stuck at a very bad local minimum.
The attacks that we present in this paper can, at least empirically, be applied in some form to non-convex models.
For example, in the \inf{} attack, we can still move poisoned data points along the gradient of the test loss.
It is an open question whether our attacks remain effective in the non-convex setting.

\paragraph{Strategies for stronger defenses.}
How might we build stronger defenses that are robust against determined attackers?
We outline several approaches, as well as potential difficulties.

One strategy would be to try to design better outlier detectors---perhaps the \Ltwo{} and \slab{} defenses
provide too crude a measure of whether a point is realistic, and sophisticated generative models such as generative adversarial nets \citep{goodfellow2014gan} could
better rule out poisoned data. We are skeptical of this approach, as there are many natural distributions (such as
multivariate Gaussians) where even a perfect generative model cannot prevent an adversary from substantially skewing
empirical statistics of the data (see \citet{steinhardt2018thesis}, Section 1.3). The existence of adversarial test
examples for image classifiers \citep{szegedy2014intriguing,goodfellow2015explaining} also weights against this
approach, since such examples are generated using a method for inducing high loss under a target model. This method could likely be
adapted for use in the \mm{} attack, as the the main sub-routine in that attack involves generating examples that induce high loss under a target model.

A different strategy is to learn multiple models on different (random or otherwise) subsets of data, in the hopes that the data will be relatively clean at least on some subsets \citep{fischler1981random,kearns1993learning,cretu2008casting}.
In general, these methods are variants of loss-based defenses, in that they assume that poisoned points tend to be poorly fit by the model, and could therefore still be vulnerable to attacks that specifically target loss-based defenses.
Especially with larger models and datasets, these methods also incur the additional computational cost from needing to fit multiple models.

Another strategy rests on the observation that if we could directly minimize the $0-1$ test error (rather than using a convex proxy),
then an adversary controlling an $\epsilon$-fraction of the data could always induce at most additional $\frac{\epsilon}{1-\epsilon}$
error, at least on the training set.\footnote{To see this, note that the $0/1$-loss of $\theta^*$ averaged across $\sDcp$ is at most
at most $\epsilon$ larger than across $\sDc$, so any $\hat{\theta}$ outperforming $\theta^*$ can only have slightly higher loss
than $\hat{\theta}$ across $\sDc$.}
The key issue with convex proxies for the $0/1$-loss is that they are unbounded and so an adversary can cause the loss on $\sDp$ to
be very large.
One could perhaps do better by using non-convex but bounded proxies for the $0/1$-loss, which would make optimization of the training loss more difficult, but might pay off with higher robustness.
However, it also opens up a new avenue for attack---the attacker could try to push the learner towards a bad local minimum.
There are also known hardness results for even approximately minimizing the
$0/1$-loss \citep{feldman2009agnostic,guruswami2009hardness},
but it is possible that they do not apply in practice.

Finally, as noted in Section~\ref{sec:related}, there is recent work seeking to design
estimators that are \emph{provably robust} to adversarial training data under suitable
distributional assumptions. \citet{diakonikolas2018sever} recently presented a practical
implementation of these methods for classification and regression tasks and showed promising initial results. Their method iteratively removes points with outlying gradients and refits the model, and can be viewed as a more sophisticated version of an iterative loss-based defense; \citet{liu2017robust} and \citet{jagielski2018manipulating} also present similar iterative algorithms for the regression setting.
We view provable security under a well-defined threat model as the gold standard, and encourage further work along this direction (see \citet{li2018thesis} or \citet{steinhardt2018thesis} for two recent overviews).
There appears to be plenty of room
both to improve the practical implementation of this family of defenses and to devise new theoretically-grounded procedures.

%% file: svm-proof.tex
\section{How many distinct points are needed for data poisoning attacks?}\label{sec:2points}

Consider some attack $\sDp$ which makes a defender learn model parameters $\hat\theta$.
Under what conditions does there exist some other attack $\sDp'$ that contains at most as many points ($|\sDp| \geq |\sDp'|$),
but with fewer \emph{distinct} points (i.e., $\sDp'$ contains repeated copies of points)?

If the attacker could place poisoned points at arbitrary locations, and if the model's loss function is unbounded (as is the case in most models, e.g., SVMs or logistic regression), then very few poisoned points (distinct or otherwise) are generally needed since the attacker can get high leverage over the model by placing a poisoned point far away.
However, in our setting, the attacker is constrained to play poisoned points that are in the feasible set $\sF$.

In this section, we provide a general method for finding the minimum number of distinct points necessary for achieving any attack, given a model with a strictly convex loss function and some feasible set $\sF$. We show that for binary SVMs and logistic regression, if $\sF$ is convex for each class---as is the case for the \Ltwo{}, \slab{}, \Loss{}, and \PCA{}---then only 2 distinct points are necessary.

As a high-level sketch, our proof proceeds as follows:
\begin{enumerate}
  \item (\refprop{car}) We consider the set of scaled feasible gradients $\sGt \eqdef \{\alpha \nabla_\theta \ell(\hat\theta; x, y): 0 \leq \alpha \leq 1, (x,y) \in \sF\}$.
  This set corresponds to the gradients of all feasible points, scaled by some $0 \leq \alpha \leq 1$. We show that the number of distinct points needed for an attack relates to the geometry of this set $\sGt$. In particular, if $\sGt$ is convex for each class, then only 2 points are needed.

  \item (\refprop{2points}) We check that for SVMs, $\sGt$ is convex for each class if the original feasible set $\sF$ is convex for each class.

  \item (\refprop{margin}) More generally, we establish conditions under which differentiable margin-based losses have $\sGt$ convex for each class, and we show that logistic regression satisfies these conditions.
\end{enumerate}
For convenience, in the sequel we will assume that these models are trained by finding
\begin{align}
\hat\theta = \underset{\theta \in \Theta}{\arg\min} \   \
\frac{\lambda}{2} \|\theta\|_2^2
+ \sum_{(x,y) \in \sDc} \ell(\theta; x, y)
+ \sum_{(\xp,\yp) \in \sDp} \ell(\theta; \xp, \yp).
\end{align}
In other words, the degree of regularization is not explicitly affected by the total number of data points $|\sDc| + |\sDp|$. Moreover, the overall loss is strictly convex due to regularization, even if $\ell(\theta; x, y)$ is only convex (and not strictly convex) in $\theta$, as is the case with the hinge loss.
We also assume that $\sDp \subseteq \sF$ (otherwise, the poisoned points will simply be thrown out by the defender).

We start by establishing the equivalence between the number of distinct points needed to poison a given model and the geometry of the set of feasible gradients of that model.

\begin{definition}
The \car{} number of a set $\sG \subseteq \R^n$ is the smallest number $c$ such that each $\tilde g \in \text{conv}(\sG)$
can be written as a convex combination of at most $c$ points in $\sG$. (Each $\tilde g$ may be a convex combination of a different set of $c$ points.)
\end{definition}

\begin{proposition}
\label{prop:car}
Consider a defender who learns a model by first discarding all points outside a fixed feasible set $\sF$,
and then finding the parameters that minimize a strictly convex loss $\ell(\theta; x, y)$ averaged over the training set.
If a parameter $\hat{\theta}$ is attainable by any set of $\tilde n$ poisoned points
$\sDp = \{(\xp_1,\yp_1),\ldots,(\xp_{\tilde n}, \yp_{\tilde n})\}  \subseteq~\sF$,
then there exists a set $\tDp$ that also attains $\hat\theta$ with at most $\tilde n$ poisoned points
but only contains $c$ distinct points (with a potentially fractional number of repeats of each point),
where $c$ is the \car{} number of the set of possible scaled gradients $\sGt \eqdef \{\alpha \nabla_\theta \ell(\hat\theta; x, y): 0 \leq \alpha \leq 1, (x,y) \in \sF\}$.

\end{proposition}

\begin{remark}
If $\ell(\theta; x, y)$ is not differentiable in $\theta$, we obtain an equivalent result by considering the subgradient sets $\partial_\theta \ell$.
In this case, we can define
$\sGt \eqdef \{\alpha g: 0 \leq \alpha \leq 1, g \in \bigcup_{(x,y) \in \sF}  \partial_\theta \ell(\hat\theta; x, y)\}$.
For clarity in the following proof, we will assume that $\ell(\theta; x, y)$ is differentiable,
but the argument for the non-differentiable case is almost identical.
\end{remark}

\begin{proof}
Assume that we are given a set of $n$ clean training points $\sDc$ and a set of $\tilde n$ poisoned points $\sDp$.
Without loss of generality, we assume that each poisoned point $(\xp, \yp) \in \sDp$ lies in the feasible set $\sF$
(otherwise, the poisoned point will be filtered out and have no effect).

The defender learns parameters $\hat\theta$ that minimize the training loss
\begin{align}
\frac{\lambda}{2} \|\theta\|_2^2
+ \sum_{(x,y) \in \sDc} \ell(\theta; x, y)
+ \sum_{(\xp,\yp) \in \sDp} \ell(\theta; \xp, \yp).
\end{align}
Since $\hat\theta$ is a minimum of the loss, we have that
\begin{align}
\label{eqn:subg}
0 = \lambda\hat\theta + \sum_{(x,y) \in \sDc} \nabla_\theta \ell(\hat\theta; x, y) +
\sum_{(\xp,\yp) \in \sDp} \nabla_\theta \ell(\hat\theta; \xp, \yp),
\end{align}
where $\nabla_\theta \ell(\hat\theta; x, y)$ denotes the gradient of the loss at the point $(x,y)$ with parameters $\hat\theta$.

Our goal is to find the minimum number of distinct points $c$ such that given any clean dataset $\sDc$, attack $\sDp$, and consequent model parameters $\hat\theta$, we can find some other attack $\tDp$ with at most $c$ distinct points such that $\tDp$ also makes the defender learn $\hat\theta$.
The key observation is that because the loss is strictly convex (by assumption), \refeqn{subg} is a necessary and sufficient condition for the defender to learn $\hat\theta$. In particular,
if we define $g \eqdef \sum_{(\xp,\yp) \in \sDp} \nabla_\theta \ell(\hat\theta; \xp, \yp)$,
then any other attack $\tDp$ that satisfies
$g = \sum_{(\xp,\yp) \in \tDp} \nabla_\theta \ell(\hat\theta; \xp, \yp)$---note that we have $\tDp$ in place of $\sDp$---
will also make the defender learn $\hat\theta$.

What values can $g$ take on?
Since $g \eqdef \sum_{(\xp,\yp) \in \sDp} \nabla_\theta \ell(\hat\theta; \xp, \yp)$ by construction,
and each point $(\xp,\yp)$ in $\sDp$ lies in the feasible set $\sF$,
we know that the normalized vector $\tilde g \eqdef g/\tilde n$ (where $\tilde n$ is the number of points in $\tDp$) is contained in the convex hull $\text{conv}(\sGt)$, where $\sGt$ is the set of scaled gradients that are possible in the feasible set, $\sGt \eqdef \{\alpha \nabla_\theta \ell(\hat\theta; x, y): 0 \leq \alpha \leq 1, (x,y) \in \sF\}$.%
\footnote{$g$ is actually contained in the convex hull of the unscaled gradient set $\{ \nabla_\theta \ell(\hat\theta; x, y): (x,y) \in \sF\}$, which is a subset of the scaled gradient set, so the above proof also goes through if we consider the unscaled gradient set in place of the scaled gradient set. However, as we will see later in this section, adding the $\alpha$ scaling term reduces the \car{} number, which gives a stronger result.}

Now, say we can find $c$ points $g_1, g_2, \ldots, g_c \in \sGt$ such that
$\tilde g$ is a convex combination of these points
(i.e., $\tilde g \in \gamma_1 g_1 + \gamma_2 g_2 + \ldots + \gamma_k g_c$, with $\gamma_i \geq 0$ and $\sum_i \gamma_i = 1$).
Since each point $g_i \in \sGt$ can be written as $\alpha_i \nabla_\theta \ell(\hat\theta; \xp_i, \yp_i)$ for some $(\xp_i, \yp_i) \in \sF$ with $0 \leq \alpha_i \leq 1$,
we can construct a dataset $\tDp$ comprising $\alpha_1 \gamma_1 \tilde n$ copies of $(\xp_1, \yp_1)$, $\alpha_2 \gamma_2 \tilde n$ copies of $(\xp_2, \yp_2)$, and so on,
such that $g = \tilde n \tilde g = \tilde n \sum_{i=1}^c \alpha_i \gamma_i \nabla_\theta \ell(\hat\theta; \xp_i, \yp_i)$. This constructed dataset $\tDp$ will therefore attain $\hat\theta$ with only $c$ distinct points, and since $\sum_{i=1}^c \alpha_i \gamma_i \leq 1$, the total weight of all of these points will be less than or equal to $\tilde n$.

We thus want to find the smallest number $c$ such that for any $\tilde g \in \text{conv}(\sGt)$,
we can write $\tilde g$ as a convex combination of at most $c$ points in $\sGt$.
This is the definition of the \car{} number of $\sGt$, as desired.

\end{proof}

\refprop{car} tells us that to find the number of distinct points required for data poisoning attacks on a given model and feasible set,
it suffices to find the \car{} number of the set of feasible gradients of that model.
Finding the \car{} number of a set is a well-studied problem (see, e.g., \citet{barany2012notes}, or \citet{mirrokni2015tight} for an approximate version of the problem).
In our setting, each feasible set can be written as the union of a small number of convex sets, which simplifies the analysis of its \car{} number.
We start by establishing the following lemma:

\begin{lemma}
\label{lem:carsets}
If a set $\sG$ is the union of $k$ convex sets, $\sG = \sG_1 \cup \sG_2 \cup \ldots \sG_k$ where each $\sG_i$ is convex,
then the \car{} number of $\sG$ is at most $k$.
\end{lemma}
\begin{proof}
Pick any $\tilde g \in \text{conv}(\sG)$. By construction, we can write $\tilde g = \sum_{i=1}^k \sum_{j=1}^{n_i} \alpha_{ij} x_{ij}$ where
$x_{ij} \in \sG_i$, $\alpha_{ij} \geq 0$, and $\sum_i \sum_j \alpha_{ij} = 1$.
Since the $\sG_i$ are convex sets, we can find $\tilde x_i \in \sG_i \subseteq \sG$ such that $\tilde x_i = \sum_{j=1}^{n_i} \alpha_{ij} x_{ij} / \sum_{j=1}^{n_i} \alpha_{ij}$, allowing us to write $\tilde g = \sum_{i=1}^k \left(\sum_{j=1}^{n_i} \alpha_{ij}\right) \tilde x_i$.
Since any $\tilde g \in \text{conv}(\sG)$ can be written as the convex combination of at most $k$ points in $\sG$, the \car{} number of $\sG$ is $k$.

\end{proof}

We use this lemma to establish the \car{} number of the set of scaled gradients for a binary SVM.

\begin{proposition}[\car{} number of $\sG$ for a 2-class SVM]
\label{prop:2points}
Consider the setting of \refprop{car}, and let the loss function be the $\ell_2$-regularized hinge loss on data with binary labels.
Suppose that for each class $y = -1, +1$, the feasible set
$\sFy \eqdef \{x: (x,y) \in \sF \}$ is a convex set.
Then the \car{} number of $\sGt \eqdef \{\alpha g: 0 \leq \alpha \leq 1, g \in \bigcup_{(x,y) \in \sF}  \partial_\theta \ell(\hat\theta; x, y)\}$ is at most 2, independent of $\hat\theta$.
\end{proposition}

\begin{proof}
Recall that in a binary SVM, the loss on an individual point is given by
\begin{align}
\ell(\theta; x, y) = \max(0, 1 - y\theta^\top x).
\end{align}
For convenience, we have folded the regularization term into the loss on each point.

From \refprop{car}, we want to find the \car{} number of the set of all possible scaled (sub)gradients of poisoned points
 $\sG \eqdef \{\alpha g: 0 \leq \alpha \leq 1, g \in \bigcup_{(x,y) \in \sF}  \partial_\theta \ell(\hat\theta; x, y)\}$.
For a binary SVM, the subgradient sets are:
\begin{align}
	\partial_\theta \ell(\hat\theta; x, y) = \left.
	\begin{cases}
		\{0\}, & \text{if } y \hat\theta^\top x > 1 \\
		\{-\gamma yx: 0 \leq \gamma \leq 1\} & \text{if } y \hat\theta^\top x = 1 \\
		\{-yx\} & \text{if } y \hat\theta^\top x < 1.
	\end{cases}
	\right.
\end{align}
Plugging this into the expression for $\sGt$, we get that
\begin{align}
\sGt = \{- \alpha yx: 0 \leq \alpha \leq 1, (x, y) \in \sF, y \hat\theta^\top x \leq 1 \},
\end{align}
which we can rewrite as the union of two convex sets, one for each class:
\begin{align}
\sGt = \{- \alpha x: 0 \leq \alpha \leq 1, (x, +1) \in \sF_{+1}, \hat\theta^\top x \leq 1 \}
\cup
\{\alpha x: 0 \leq \alpha \leq 1, (x, -1) \in \sF_{-1}, -\hat\theta^\top x \leq 1 \}. \nonumber
\end{align}
By \reflem{carsets}, the \car{} number of $\sGt$ is at most 2, regardless of what $\hat\theta$ is.

\end{proof}

More generally, we can bound the \car{} number of the set of scaled gradients for a particular class of convex, differentiable, margin-based losses.
\begin{proposition}[\car{} number of $\sG$ for margin-based losses]
\label{prop:margin}
Consider the setting of \refprop{car} on binary data, and let the loss function be $\ell(\theta; x, y) = c(-y\theta^\top x)$, where $c: \R \to \R$ is a convex, monotone increasing, and twice-differentiable function.
Suppose that the ratio of the second to the first derivative of $c$, $c'' / c'$ is a monotone non-increasing function,
and that for each class $y = -1, +1$, the feasible set
$\sFy \eqdef \{x: (x,y) \in \sF \}$ is a convex set.
Then the \car{} number of $\sGt \eqdef \{\alpha \nabla_\theta \ell(\hat\theta; x, y): 0 \leq \alpha \leq 1, (x,y) \in \sF\}$ is at most 2, independent of $\hat\theta$.

\end{proposition}
\begin{proof}
We can write $\sGt$ as the union of two sets $\sG_{\hat\theta, +1}$ and $\sG_{\hat\theta, -1}$, one for each class,
where $\sGty \eqdef \{\alpha \nabla_\theta \ell(\hat\theta; x, y): 0 \leq \alpha \leq 1, x \in \sFy\}$.
To show that the \car{} number of $\sGt$ is at most 2, it suffices to show that each class set $\sGty$ is convex and then apply \reflem{carsets}.

Pick $\hat\theta$ arbitrarily and fix $y$ to be $-1$ or $+1$.
To check that $\sGty$ is convex, it suffices to check that
for each possible choice of $x_1$ and $x_2$ in $\sFy$ and $0 \leq \gamma \leq 1$,
there exists some $\xt \in \sFy$ and $0 \leq \alpha \leq 1$ such that
\begin{equation}\label{eqn:xt-unexpanded}
\alpha \nabla_\theta \ell(\hat\theta; \xt, y) = \gamma \nabla_\theta \ell(\hat\theta; x_1, y) + (1 - \gamma) \nabla_\theta \ell(\hat\theta; x_2, y).
\end{equation}
Since our loss function has the form $\ell(\theta; x, y) = c(-y\theta^\top x)$, we have that
\begin{equation}
\nabla_\theta \ell(\hat\theta; x, y) = c'(-y\theta^\top x) (-yx),
\end{equation}
where $c'$ is the derivative of $c$.
Substituting this into \refeqn{xt-unexpanded} and cancelling out the $-y$ terms on both sides gives us the equivalent condition
\begin{equation}\label{eqn:convex-condition}
\alpha c'(-y\theta^\top \xt) \xt = \gamma c'(-y\theta^\top x_1) x_1 + (1 - \gamma) c'(-y\theta^\top x_2) x_2.
\end{equation}

To satisfy the above condition, we will take
\begin{align}\label{eqn:xt-expanded}
\xt = \frac{\gamma c'(-y\theta^\top x_1)}{\gamma c'(-y\theta^\top x_1) + (1 - \gamma) c'(-y\theta^\top x_2)}x_1
+ \frac{(1 - \gamma) c'(-y\theta^\top x_2)}{\gamma c'(-y\theta^\top x_1) + (1 - \gamma) c'(-y\theta^\top x_2)}x_2,
\end{align}
which in turn implies that
\begin{align}\label{eqn:alpha-expanded}
\alpha = \frac{\gamma c'(-y\theta^\top x_1) + (1 - \gamma) c'(-y\theta^\top x_2)}{c'(-y\theta^\top \xt)}.
\end{align}
Since $\xt$ is a convex combination of $x_1$ and $x_2$, the convexity of $\sFy$ implies that
$\xt \in \sFy$, so it remains to check that $0 \leq \alpha \leq 1$.\footnote{
Note that since $\alpha c'(-y\theta^\top \xt)$ is a scalar, so for \refeqn{convex-condition} to hold, $\xt$ needs to point in the same direction as
$\gamma c'(-y\theta^\top x_1) x_1 + (1 - \gamma) c'(-y\theta^\top x_2) x_2$.
Moreover, since we need $\xt$ to be in $\sFy$, we want $\xt$ to be a convex combination of $x_1$ and $x_2$.
This choice of $\xt$ is the only choice that satisfies these two considerations.
}
Moreover, since $c$ is monotone increasing by assumption, $c'$ is positive, and therefore $\alpha$ is positive.
It remains to check that $\alpha \leq 1$.

First, note that $\log c'(\cdot)$ is a concave function,
as its derivative $c'' / c'$ is monotone non-increasing by assumption.
For notational convenience, let $s_1 \eqdef -y \theta^\top x_1$ and $s_2 \eqdef -y \theta^\top x_2$,
and let $T \eqdef \gamma c'(s_1) + (1 - \gamma) c'(s_2)$.
We then have that
\begin{align*}
\log c'(-y\theta^\top \xt) &=
\log c'\left(\frac{\gamma c'(s_1)s_1 + (1 - \gamma)c'(s_2)s_2}{T} \right)\\
&\geq (\gamma c'(s_1) / T) \log c'(s_1) + ((1 - \gamma)c'(s_2) / T) \log c'(s_2) \hspace{16mm}\text{(Jensen's)}\\
&= (\gamma c'(s_1) / T) \log \frac{\gamma c'(s_1) / T}{\gamma} + ((1 - \gamma)c'(s_2) / T) \log \frac{(1 - \gamma) c'(s_2) / T}{\gamma} + \log T\\
&\geq \log T \hspace{44mm}\text{(non-negativity of KL divergence)}
\end{align*}
Exponentiating both sides and rearranging gives us
\begin{align*}
\alpha &= \frac{T}{c'(-y\theta^\top \xt)} \leq 1.
\end{align*}

The above argument shows that $\sGty$ is a convex set. Since we picked $\hat\theta$ and $y$ arbitrarily, we can apply \reflem{carsets} to conclude that $\sGt = \sG_{\hat\theta, +1} \cup \sG_{\hat\theta, -1}$ has \car{} number at most 2 regardless of $\hat\theta$.
\end{proof}

\begin{corollary}[\car{} number of $\sG$ for logistic regression]\label{cor:logistic}
Consider a logistic regression model in the setting of \refprop{margin},
where $\ell(\theta; x, y) = \log(1 + \exp(-y\theta^\top x))$.
Then the \car{} number of $\sGt \eqdef \{\alpha \nabla_\theta \ell(\hat\theta; x, y): 0 \leq \alpha \leq 1, (x,y) \in \sF\}$ is at most 2, independent of $\hat\theta$.
\end{corollary}
\begin{proof}
In logistic regression, we have that $c(v) = \log(1 + \exp(v))$; $c'(v) = \sigma(v)$ where $\sigma$ is the sigmoid function, $\sigma(v) \eqdef \frac{1}{1+\exp(-v)}$;
and $c''(v) = \sigma(v) (1 - \sigma(v))$.
Thus, $\frac{c''}{c'}$ is a monotone decreasing function, and $c$ is convex, monotone increasing, and twice-differentiable,
so \refprop{margin} applies.
\end{proof}

We can collect all of the above results into the following theorem, which appears in the main text.

\begingroup
\def\thetheorem{\ref{thm:2points}}
\begin{theorem}[2 points suffice for 2-class SVMs and logistic regression]
Consider a defender that learns a 2-class SVM or logistic regression model by first discarding all points outside a fixed feasible set $\sF$ and then minimizing the average (regularized) training loss. Suppose that for each class $y = -1, +1$, the feasible set
$\sFy \eqdef \{x: (x,y) \in \sF \}$ is a convex set.
If a parameter $\hat{\theta}$ is attainable by any set of $\tilde n$ poisoned points
$\sDp = \{(\xp_1,\yp_1),\ldots,(\xp_{\tilde n}, \yp_{\tilde n})\}  \subseteq~\sF$,
then there exists a set of at most $\tilde n$ poisoned points $\tDp$ (possibly with fractional copies)
that also attains $\hat\theta$ but only contains $2$ distinct points, one from each class.

More generally, the above statement is true for any margin-based model with loss of the form $\ell(\theta; x, y) = c(-y\theta^\top x)$,
where $c: \R \to \R$ is a convex, monotone increasing, and twice-differentiable function, and the ratio of second to first derivatives $c''/c'$ is monotone non-increasing.
\end{theorem}
\addtocounter{theorem}{-1}
\endgroup
\begin{proof}
From \refprop{2points} (SVMs), \refprop{margin} (margin-based losses), and \refcor{logistic} (logistic regression), we have that the \car{} number of $\sGt$ (the set of scaled possible gradients) for each of these models is at most 2, regardless of $\hat\theta$. By \refprop{car}, we conclude that only 2 distinct points are necessary.
In particular, since $\sGt$ can be represented in each of these cases as the union of two convex sets, one for each class, we need 1 distinct point from each class to realize any data poisoning attack.
\end{proof}

The general approach of finding the \car{} number of the set of scaled possible gradient can be applied to other models beyond those that we consider in this paper.
As one example, we can extend the above approach to the setting of a multi-class SVM:

\begin{proposition}[$k(k-1)$ distinct points suffice for a $k$-class SVM]
\label{prop:kpoints}
Consider the setting of \refprop{2points}, but with a $k$-class SVM.
If a parameter $\hat{\theta}$ is attainable by any set of $\tilde n$ poisoned points
$\sDp = \{(\xp_1,\yp_1),\ldots,(\xp_{\tilde n}, \yp_{\tilde n})\}  \subseteq~\sF$,
then there exists a set of at most $\tilde n$ poisoned points $\tDp$
that also attains $\hat\theta$, but that only has $k-1$ distinct values of $\xp$ for each distinct $\yp$,
for a total of $k(k-1)$ distinct points.

\end{proposition}
\begin{proof}
We follow the multi-class SVM formulation described in \citet{crammer2002learnability},
which has parameters $\theta = [\theta_1; \theta_2; \ldots; \theta_k] \in \R^{kd}$ for each class $y = 1, 2, \ldots, k$,
where $d$ is the dimensionality of the feature space $\sX$.
The loss on an individual point is given by
\begin{align}
\ell(\theta; x, y) = \max(0, 1 - \theta_y^\top x + \max_{i \neq y}\theta_i^\top x).
\end{align}
This reduces to the above formulation for a binary (2-class) SVM by setting $\theta_{-1} = -\theta_{+1}$.

Let $i_{x,y} = \arg\max_{i \neq y}\theta_i^\top x$, and let
$x^{(i)} = [\underbrace{0, \ldots, 0,}_{(i-1)d \text{ zeroes}} x, \underbrace{0, \ldots, 0}_{(n-i)d \text{ zeroes}}]$.
Without loss of generality, we can ignore subgradients and take
$\nabla_\theta \ell(\theta; x, y) = -x^{(y)} + x^{(i_{x,y})}$,
following a similar argument to that used in \refprop{2points}.

Define $\sG_{i, j} = \{-\alpha x^{(i)} + \alpha x^{(j)} : 0 \leq \alpha \leq 1, (x, i) \in \sF \}$. We have that $\sGt = \bigcup_{i=1}^k \bigcup^k_{j \neq i} \sG_{i, j}$,
and since each $\sG_{i, j}$ is a convex set, by \reflem{carsets}, the \car{} number of $\sG$ is at most $k(k-1)$.
\end{proof}

%% file: app_attack_details.tex
\section{Attack implementation details}\label{sec:app_attack_details}

\subsection{The \inf{} attack}\label{sec:app_inf_details}

The following details apply to both the \infbasic{} and \inf{} attacks.

\textbf{Smoothed hinge loss.} In our setting, the loss $\ell(\theta; x,y)$ is the hinge loss
$\max(0, 1 - y\theta^{\top}x)$.
One issue is that this loss is piecewise linear, which means that its gradient is zero or one, and its Hessian
zero, almost everywhere.
As a result, the gradient of the test loss $L$ w.r.t. $\xp$---which involves an inverse-Hessian-gradient-product, as in \refeqn{influence-params}---gives a poor indication of which directions to
perturb $\xp$, and can easily get stuck.
To mitigate this problem, we follow \citet{koh2017understanding} and smooth the hinge loss for the purposes of computing the gradient
(the actual training of the model is still done using the hinge loss). Specifically, we replace it with the function
$\ell_{\text{smooth}}(\theta; x,y) = \delta \log(1 + \exp(\frac{1-y\theta^{\top}x}{\delta}))$
for some small $\delta$. The function $\ell_{\text{smooth}}$ converges to
$\ell$ as $\delta \to 0$, but has derivatives of all orders whenever $\delta > 0$.
We note that there are other ways to get around the non-differentiability of the hinge loss. For example,
\citet{biggio2012poisoning},
\citet{mei2015teaching}, and their subsequent work
make use of the dual formulation of SVMs to derive the required gradient,
under the assumption that the set of support vectors (defined as training points that are exactly at the hinge) remain unchanged by the gradient update.
For our purposes, we expect that such methods would perform similarly if the step size of the gradient update is small enough.

\textbf{Computing inverse Hessian-vector products.}
To efficiently compute $\frac{\partial \hat{\theta}}{\partial \xp}$ in \refeqn{influence-params},
we follow \citet{koh2017understanding} and use a combination of fast Hessian-vector products
\citep{pearlmutter1994fast} and a conjugate gradient solver \citep{martens2010deep};
see also \citet{agarwal2016second} and \citet{munoz2017towards} for other tractable approaches.
We select the step size $\eta$ by trying a range of options and selecting the best-performing one on the test set
(since, in our setting, the attacker knows the test set in advance).

\textbf{Choosing the labels of poisoned points.}
For the \infbasic{} attack, we initialize both $\xt_i$ and $\yt_i$ through random label flips on the training set. In other words, we select $\epsilon n$ data points from the clean data $\sDc$ uniformly at random, with replacement, to flip and add to the poisoned data $\sDp$. We only consider data points that would still lie in $\sF_\beta$ after the label flip.
For the \inf{} attack, there is only one distinct positive poisoned point $\xpp$ and one distinct negative poisoned point $\xpn$. We weight these two points inversely proportionally to the class balance: i.e., if there are $P$ positive and $N$ negative points in $\sDc$, then we place $\frac{N}{P+N} \cdot \epsilon$ weight on $(\xpp, 1)$
and $\frac{P}{P+N} \cdot \epsilon$ weight on $(\xpn, {-1})$.
This maintains the same class balance, on average, as the label flip initialization for the \infbasic{} attack.

\subsection{The \kkt{} attack}\label{sec:app_kkt_details}

\textbf{Generating decoy parameters.} For each dataset, we first generated candidate decoy parameters as in \refsec{kkt-choosing-decoys}---adding $r$ copies
of each test point whose flipped label has loss greater than $\gamma$. For Enron, we swept over $r \in \{1,2,3,5,8,12,18,25,33\}$ and
$\gamma$ set to the $q$th quantile of the loss (over the flipped test set), for $q \in \{0.05, 0.10, \ldots, 0.55\}$.
This yielded $99$ candidates $\thetadecoy$; for efficiency we removed all parameters that had lower test error and higher training loss
than some other candidate $\thetadecoy'$. This left us with $48$ parameters total. For IMDB, we applied a similar procedure but took
$r \in \{1,2,3,4,5\}$ and $q \in \{0.1,0.2,\ldots,0.6\}$; this yielded $18$ candidates after pruning (we sought fewer candidates for
IMDB because it is bigger and slower to attack).

\textbf{Choosing the labels of poisoned points.} Given the fraction of poisoned data $\epsilon = 3\%$, for each set of decoy parameters, we grid searched over
7 different ratios of positive vs. negative poisoned points,
ranging from $\epp = 3\%, \epn = 0\%$ to $\epp = 0\%, \epn = 3\%$.

\subsection{The \mm{} attack}\label{sec:app_mm_details}
\textbf{Loss defense threshold.} To evade the \Loss{} defense, we used the threshold $\tau = 0.25$ across all experiments. The results were fairly robust to this choice of threshold.

\textbf{Multi-class training.} For the MNIST dataset, we trained the multi-class SVM  with AdaGrad \citep{duchi10adagrad}, with a batch size of 20, step size $\eta = 0.02$, and 3 passes over the training data.

%% file: experiments_mnist_dogfish.tex
\section{Experiments on \mnist{} and Dogfish}\label{sec:experiments-mnist-dogfish}

In this section, we study the effectiveness of the \inf{} attack on the following two datasets:
\begin{enumerate}
\item The \mnist{} image dataset \citep{lecun1998gradient}, which requires each input feature to lie within the interval [0, 1] (representing normalized pixels). It is derived from the standard 10-class MNIST dataset by taking just the images labeled `1' or `7'.
It is easily linearly separable: an SVM achieves 0.7\% error on the clean data.
\item The Dogfish image dataset \citep{koh2017understanding}, which has no input constraints; its features are neural network representations,
which for the purposes of this paper we allow to take any value in $\R^d$.
Compared to the \mnist{} dataset (where $n \gg d$), the Dogfish dataset has $n \approx d$,
allowing attackers to potentially exploit overfitting. The Dogfish dataset also has a low base error of 1.3\%.
\end{enumerate}

\begin{figure}[h]
\begin{center}
\includegraphics[width=\columnwidth]{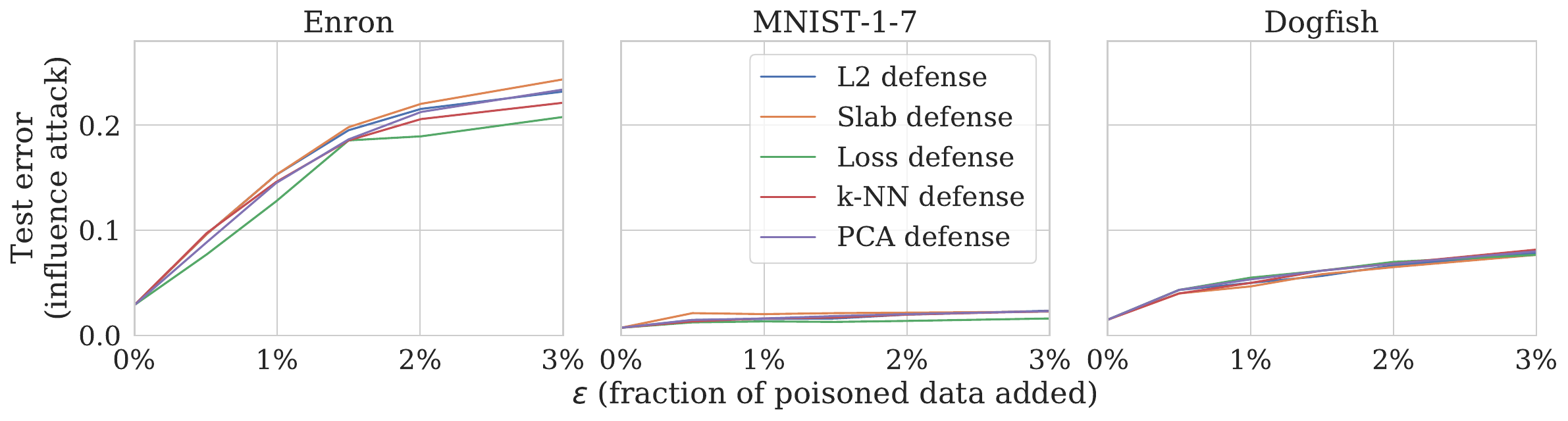}
\end{center}
\vspace{-5mm}
\caption{The \inf{} attack successfully drives test error on the Enron dataset
from 3\% to 23\% with just $\epsilon = 3\%$ poisoned data, and on the Dogfish dataset from 1\% to 8\%, but does not manage to significantly affect the \mnist{} dataset.}
\label{fig:influence}
\end{figure}

As discussed in \refsec{experiments}, \citet{steinhardt2017certified} showed that the \Ltwo{} and \slab{} defenses are certifiably effective at defending the \mnist{} dataset, and to a lesser extent the Dogfish dataset, for low values of $\epsilon$.
The results in \reffig{influence} are consistent with this: with $\epsilon=3\%$,
the \inf{} attack increases Dogfish test error from 1\% to 8\% and does not appreciably affect \mnist{} test error. In contrast, it drives test error on the Enron dataset from 3\% to 23\%.

%% file: alfa-details.tex
\section{Label Flip Attacks}
\label{sec:alfa}

Label flip attacks are a popular strategy in the literature for executing data poisoning attacks
\citep{biggio2011label, xiao2012adversarial, xiao2015contamination}.
In a label flip attack, the attacker is given a set of real data $\sDho = \{(\xt_i, \yt_i)\}_{i=1}^{\nho}$.
To form the poisoned data $\sDp$, the attacker chooses $\epsilon n$ points (possibly repeated) from $\sDho$ and flips their labels.

Many ways of choosing which points to flip have been proposed
(see \citet{xiao2015contamination} for a review).
Here, we consider the \alfa{} attack from \citet{xiao2012adversarial}.
\alfa{} seeks to add points that have a high loss under the original (clean) model $\theta^* \eqdef \argmin_{\theta} L(\theta; \sDc)$
but a low loss under the final (poisoned) model $\hat\theta \eqdef \argmin_{\theta} L(\theta; \sDcp)$.
Concretely, it solves:
\begin{align}
\nonumber \underset{\sDp}{\text{maximize}} \quad
& L(\hat\theta; \sDp) - L(\theta^*; \sDp) \\
\text{s.t.} \quad
& |\sDp| = \epsilon |\sDc|\nonumber\\
& (\xt, -\yt) \in \sDho \quad \forall (\xt, \yt) \in \sDp\nonumber\\
& \hat{\theta} = \argmin_{\theta} L(\theta; \sDcp) \nonumber
\end{align}
We adapt the original \alfa{} attack to our setting in the following two ways:
\begin{enumerate}
  \item We constrain all poisoned points in $\sDp$ to lie in the feasible set $\sFb$.
  \item We set $\sDho = \sDtest$; that is, the attacker gets to add points from the flipped test set. Intuitively, adding flipped versions of test points to the training set should cause the model to
  wrongly classify those test points, in line with the attacker's goal of increasing the model's loss on the test set.
\end{enumerate}

We evaluated this variant of \alfa{} on the Enron dataset.
To make it easier for the attacker, we only considered the \Ltwo{} defense.
Our implementation of the \alfa{} attack was not able to increase the test error much, achieving a $2\%$ increase in test error with $\epsilon=3\%$.
In contrast, the attacks we introduce in the main text achieve an increase in test error of $15\%$ to $20\%$, despite having to deal with more sophisticated defenses.
Our conclusion is that only modifying the labels $\yt$, as in label flip attacks,
does not give the attacker much power relative to being able to modify $\xt$ as well.

\begin{figure}[h]
\begin{center}
\includegraphics[width=0.45\textwidth]{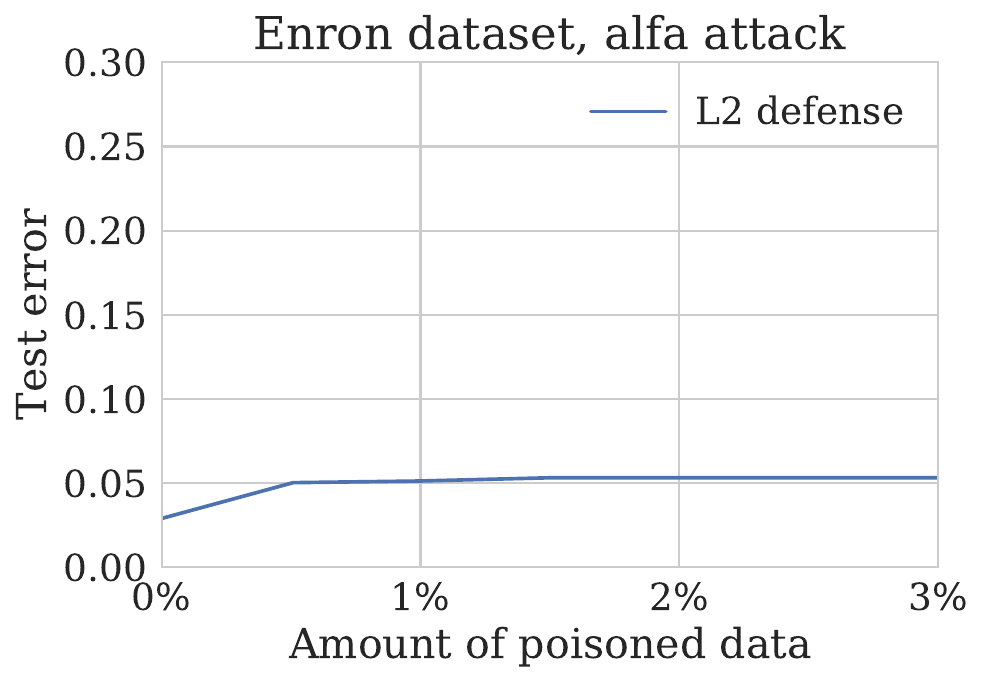}
\vspace{-3mm}
\caption{Results of the \alfa{} attack on the Enron dataset and \Ltwo{} defense.}
\end{center}
\end{figure}